\documentclass{article}
\usepackage[T1]{fontenc}
\usepackage{iclr2027_conference,times}
\usepackage{float}
\usepackage{booktabs}
\usepackage{multirow}
\usepackage{xcolor}
\usepackage{hyperref}
\usepackage{url}
\usepackage{graphicx}
\usepackage{wrapfig}
\usepackage{amsmath,amssymb}

\title{Bison: Cross-Dataset Learning for Unseen-Compound Perturbation Prediction}
\iclrfinalcopy
\author{%
  \begin{minipage}[t]{\dimexpr\textwidth-2\tabcolsep\relax}
  \centering\bfseries
  Yunfan Liu\textsuperscript{1}\quad
  Kasra Ghorbani\textsuperscript{2}\quad
  Yufei Huang\textsuperscript{2}\\
  Zicheng Liu\textsuperscript{3}\quad
  Jiangbin Zheng\textsuperscript{2}\quad
  Jingbo Zhou\textsuperscript{2}\\
  Shaorong Chen\textsuperscript{2}\quad
  Chang Yu\textsuperscript{4}\quad
  Stan Z. Li\textsuperscript{2}\\[0.6em]
  {\normalfont
  \textsuperscript{1}\,Zhejiang University\quad
  \textsuperscript{2}\,Westlake University\\
  \textsuperscript{3}\,Beihang University\quad
  \textsuperscript{4}\,Nanjing University}
  \end{minipage}%
}
\hypersetup{
  hidelinks,
  pdftitle={Bison: Cross-Dataset Learning for Unseen-Compound Perturbation Prediction},
  pdfauthor={Yunfan Liu, Kasra Ghorbani, Yufei Huang, Zicheng Liu, Jiangbin Zheng, Jingbo Zhou, Shaorong Chen, Chang Yu, Stan Z. Li}
}

\begin{document}
\maketitle
\fancyhead{}
\renewcommand{\headrulewidth}{0pt}
\suppressfloats[t] 
\begin{abstract}
Predicting transcriptional responses to unseen compounds is limited by fragmented chemical coverage and heterogeneous experimental platforms and gene panels. To assess molecular generalization across these settings, we build on Chem-PerturBridge to benchmark eight datasets with 16,771 compounds, withholding test compounds from every training dataset. This comparison reveals that high overall response agreement can coexist with weak prediction of drug-specific differences, despite reproducible signals across repeated measurements. To exploit complementary chemical supervision while targeting these differences, we introduce Bison: a shared gene representation connects native panels, while two discrete diffusion models compose context-dependent responses with molecular deviations learned through matched drug-contrast supervision. A single Bison model jointly trained across all eight datasets achieves the highest mean overall-response and drug-contrast Pearson correlations on the full benchmark in comparison with 11 methods trained independently per dataset. Compared with dataset-specific training of the same architecture, joint training increases mean drug-contrast correlation by 27.4\%, with gains across all eight datasets and improvements in overall response prediction. These results demonstrate how matched drug contrasts turn complementary screens into shared molecular supervision for unseen-drug response prediction while preserving native gene measurements.

\end{abstract}
\makeatletter
\let\bisonoriginalsection\section
\let\bisonoriginalsubsection\subsection
\renewcommand\section{\@startsection{section}{1}{\z@}{-1.0ex plus -.2ex minus -.1ex}{0.6ex plus .1ex}{\large\sc\raggedright}}
\renewcommand\subsection{\@startsection{subsection}{2}{\z@}{-.8ex plus -.2ex minus -.1ex}{0.4ex plus .1ex}{\normalsize\sc\raggedright}}
\makeatother
\section{Introduction}
\label{sec:intro}
\begingroup
\setlength{\textfloatsep}{12pt}
\begin{figure}[t]
    \centering
    \vspace{-4pt}
    \setlength{\abovecaptionskip}{3pt}
    \setlength{\belowcaptionskip}{-6pt}
    \includegraphics[width=\textwidth]{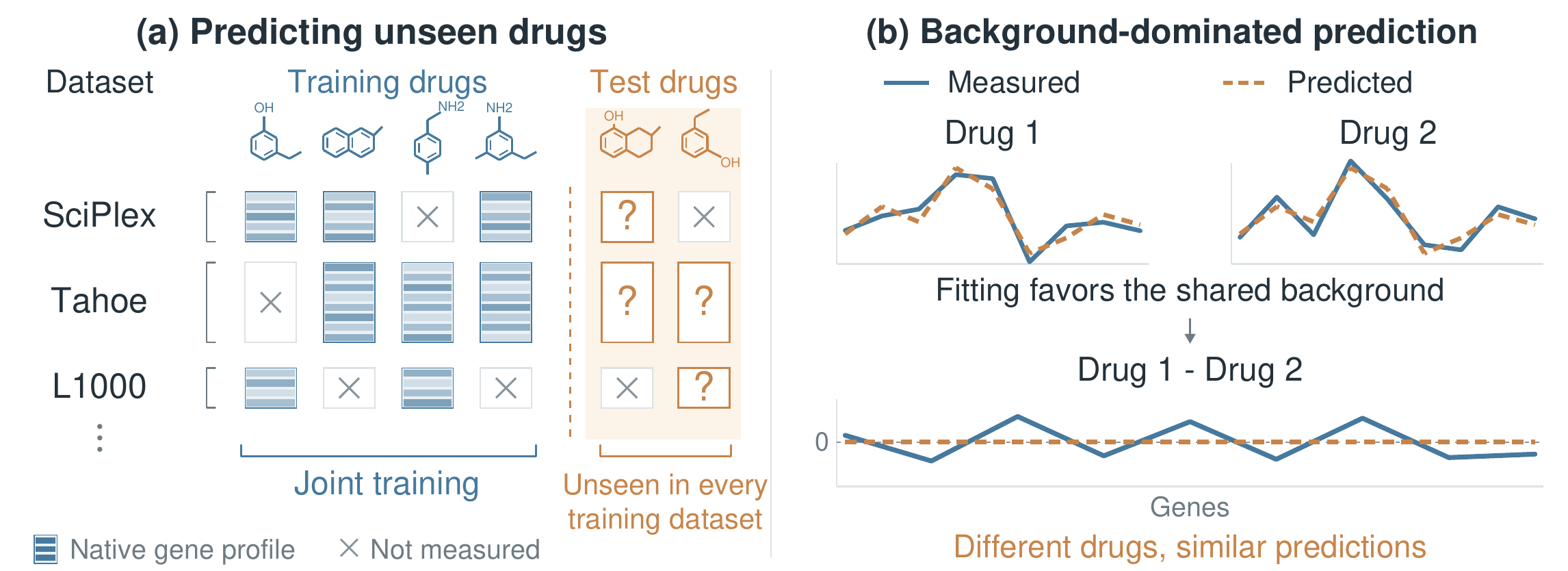}
    \caption{\textbf{The task and the evaluation challenge.} \textbf{(a)} Joint training combines observed drug-response profiles across datasets with different native gene panels. Test drugs are absent from every training dataset; crosses denote unmeasured profiles. \textbf{(b)} The shared background dominates overall responses and is easier to fit than drug-specific variation. Fitting this common component pulls predictions for different compounds toward the same profile, suppressing drug-specific differences.}
    \label{fig:overview}
\end{figure}

Predicting transcriptional responses to chemical perturbations links molecular structure to cellular function. Such predictions can help interpret drug mechanisms, prioritize compounds for experimental screening, and identify potential adverse effects~\citep{qi2024prnet}. Extending this capability to unmeasured molecules would allow limited experimental resources to explore a much larger chemical space. However, obtaining the data needed for this goal remains difficult: screening diverse compounds across cellular contexts, doses, and timepoints is costly, and a large number of profiled cells need not imply broad chemical coverage. Available data are consequently spread across studies with limited chemical coverage and heterogeneous assays, controls, and gene panels~\citep{szalata2026chemperturbridge}. This creates a coupled difficulty: individual datasets offer limited supervision for unseen molecules, while differences between experiments impede its reuse across studies.

Existing models apply molecular representations with conditional generation to predict responses to unseen drugs~\citep{hetzel2022chemcpa,qi2024prnet}, but sharing chemical supervision across datasets also requires bridging heterogeneous experiments and gene panels.
LPM advances such cross-experiment learning, yet its original identity-based perturbation embeddings do not directly support drugs absent from the entire training collection~\citep{miladinovic2025lpm}.
These advances motivate a common evaluation of generalization to globally unseen compounds across heterogeneous experimental settings.
Chem-PerturBridge provides a harmonized data foundation for this comparison~\citep{szalata2026chemperturbridge}.
Building on this resource, we benchmark 11 methods trained independently per dataset and native gene view across eight datasets with 16,771 compounds, using global compound holdout and consistent control-matching and scoring rules.

This comparison reveals that strong overall response agreement does not ensure accurate prediction of differences between compounds.
Within-plate drug contrasts expose this gap, while repeated measurements across multiple datasets show reproducible signal that existing models do not fully capture.
Meanwhile, 89.6\% of benchmark compounds occur in only one dataset, so training on individual screens leaves substantial complementary chemical supervision unused.
Together, these findings motivate joint learning that shares molecular supervision across datasets and explicitly targets differences between compounds under matched experimental conditions (Figure~\ref{fig:overview}).

To use this complementary supervision for unseen-drug prediction, we introduce Bison, which connects native gene panels through shared gene representations in a common discrete latent space.
Within this space, two discrete diffusion models (DLMs) learn across datasets with complementary objectives.
The context-response model learns complete conditional profiles, while the molecular-effect model additionally targets differences between compounds through matched drug-contrast supervision.
At inference, a shared gene-query decoder restores predictions to their native panels, where Bison combines the context branch's reference response with the molecular branch's drug-versus-reference deviation to obtain the final profile.

A single jointly trained Bison model serves all eight datasets and achieves the highest mean overall-response and drug-contrast Pearson correlations on the full benchmark in comparison with 11 methods trained independently per dataset.
With the architecture held fixed, joint training improves both metrics over dataset-specific training, increasing mean drug-contrast correlation by 27.4\%, with gains across all eight datasets.
Our contributions are:
\begin{itemize}
    \item \textbf{Revealing a molecular generalization gap.}
    Through global compound holdout and matched drug contrasts, we show that strong overall-response agreement can coexist with weak prediction of reproducible drug-specific differences.
    \item \textbf{Cross-dataset molecular learning.}
    Bison connects native gene panels through shared representations and learns molecular deviations through matched drug-contrast supervision, composing them with context-dependent responses to predict unseen-drug effects.
    \item \textbf{Gains from joint learning.}
    With the architecture held fixed, joint training improves drug-contrast prediction over dataset-specific training on all eight datasets. Case studies show that close analogs from other screens can improve predictions for unseen compounds.
\end{itemize}
\endgroup

\section{Related Work}
\label{sec:related}

\paragraph{Chemical perturbation models.} Chemical perturbation models predict transcriptional responses from molecular information and experimental context. A simple reference is ridge regression on molecular fingerprints~\citep{hoerl1970ridge,rogers2010ecfp}. Neural approaches include conditional generative and disentanglement models~\citep{hetzel2022chemcpa,piran2024biolord,qi2024prnet}, flow- and transport-based models~\citep{yu2025perturbnet,klein2025cellflow,driessen2026cmonge}, and attention-based models that incorporate cellular representations or biological priors~\citep{ji2024prophet,guo2026xpert,alsulami2026preprct}. Recent frameworks such as LPM and State further emphasize learning across heterogeneous experiments and cellular contexts~\citep{miladinovic2025lpm,adduri2025state}. We study how joint learning across complementary chemical screens improves population-level response prediction for globally unseen compounds.

\paragraph{Heterogeneous gene panels and perturbation benchmarks.} Learning across datasets requires flexible gene interfaces and consistent evaluation. Gene-aware and set-based models accommodate different gene subsets~\citep{cui2024scgpt,palla2026scldm}, while latent-code approaches separate expression representation from conditional generation~\citep{bhattacharya2026evae}. OP3 and PerturBench establish shared prediction tasks and evaluation frameworks~\citep{szalata2024openproblems,wu2025perturbench}. Chem-PerturBridge harmonizes bulk and pseudobulk profiles across assays and examines cross-dataset agreement and pooled training~\citep{szalata2026chemperturbridge}. Systema and TxPert further highlight how systematic variation and experimental references affect evaluation of genetic perturbation predictions~\citep{vinas2025systema,wenkel2026txpert}. We bring these concerns together in a chemical perturbation benchmark that retains native gene panels and evaluates both overall responses and drug-specific differences on globally unseen compounds.

\begingroup
\setlength{\intextsep}{6pt plus 2pt minus 1pt}
\setlength{\textfloatsep}{8pt plus 2pt minus 1pt}
\section{Benchmarking Generalization to Unseen Compounds}\label{sec:benchmark}

\subsection{Data and Prediction Protocol}\label{sec:benchmark_protocol}

We evaluate population-level responses using Chem-PerturBridge's harmonized bulk and pseudobulk profiles~\citep{szalata2026chemperturbridge}. The benchmark retains replicate records across eight datasets, 16,771 treated compounds and 131 cellular contexts. Figure~\ref{fig:chemical_space} shows their chemical space. Six datasets provide Full and 2,000-gene HVG panels, consistent in size with the Seurat v3 workflow~\citep{stuart2019integration}; the two L1000 datasets each provide 978 landmark genes, giving 14 views. Full retains measured genes within the benchmark vocabulary; HVG uses released variance panels selected from all chemically eligible profiles, including held-out profiles. Panels specify input and output genes (Appendix~\ref{app:data_coverage}).

\begin{figure}[H]
\centering
\vspace{-7pt}
\includegraphics[width=0.98\textwidth]{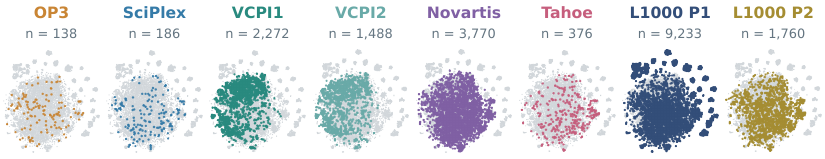}\par
\vspace{-7pt}
\caption{\textbf{Chemical space across datasets.} Shared Morgan-fingerprint t-SNE of 16,771 compounds. Colors highlight each dataset against the gray background; $n$ denotes compound counts.}
\label{fig:chemical_space}
\vspace{-2pt}
\end{figure}

Compounds are globally partitioned by standardized structure into TRAIN/VAL/TEST at approximately 70\%/10\%/20\%; all treated records follow their compound's assignment. We conduct experiments on three independent global compound splits. TRAIN controls and responses to other TRAIN compounds remain available in the target experimental settings. Eleven baseline families are fitted separately per dataset and view on these aggregate profiles, then evaluated on the same held-out records; Linear and Global Mean provide additional controls.

\subsection{Evaluation Metrics}\label{sec:benchmark_metrics}

\textbf{Plate-matched overall response.}
Following \citet{wenkel2026txpert}, we evaluate overall responses with \textbf{plate-matched Pearson $\Delta$}.
For each observed profile and its prediction, we subtract the mean TRAIN control profile matched by cellular context, plate, and time.
We average these deltas over biological replicates within each compound--context--dose--time condition, compute Pearson correlation across genes, and average over conditions.
This scoring reference is shared by all methods.

\textbf{Drug-contrast correlation.}
Even with plate-matched controls, responses shared across compounds can dominate overall correlation: a model may score well while predicting nearly identical effects for different drugs.
Moreover, subtracting the same estimated control introduces a shared reference error into observed and predicted deltas~\citep{nicol2026spurious}.
To directly evaluate differences between compounds under matched experimental conditions, we introduce \textbf{drug-contrast correlation}.
Let $S$ contain at least three compounds measured in the same context, plate, dose, and time, and let $\boldsymbol\delta_{Si}$ and $\widehat{\boldsymbol\delta}_{Si}$ denote the observed and predicted replicate-averaged responses of compound $i$.
We center observed and predicted responses separately across compounds and correlate the resulting contrasts across genes:
\begin{equation}
R_{\mathrm{contrast}}
=\underset{(S,i)\in\mathcal V}{\operatorname{mean}}
\rho_{\mathrm{genes}}\!\left(
\widehat{\boldsymbol\delta}_{Si}-\overline{\widehat{\boldsymbol\delta}}_S,
\boldsymbol\delta_{Si}-\overline{\boldsymbol\delta}_S
\right),
\label{eq:drug_contrast_score}
\end{equation}
where bars denote averages across compounds in $S$, and $\mathcal V$ contains group--compound pairs with defined correlations.
Centering cancels additive components shared within each group, including the common control reference, so the score measures agreement in differences between drugs.
Predictions constant across compounds yield zero contrast vectors and undefined correlation.

\textbf{Repeatability reference.}
We assess repeatability by correlating the measured drug contrasts of the same compound under matched conditions.
These measurements come from different plates, or from replicate wells for VCPI1/2.
We report the model's mean drug-contrast correlation as a percentage of the mean repeat correlation on the same samples.
Matching and averaging details are provided in Appendix~\ref{app:repeatability}.

\textbf{Complementary metrics.}
Cell-Eval adds MAE/MSE $\Delta$ for magnitude errors, perturbation discrimination for condition identification, and DE direction agreement and overlap@100 for affected-gene recovery.
Correlations cover fourteen native views; Cell-Eval covers twelve count-based views.

\subsection{Overall Fit Can Miss Drug Differences}\label{sec:benchmark_findings}

\begin{figure}[!htb]
\centering
\vspace{-4pt}
\includegraphics[width=0.92\textwidth,trim=0 3bp 0 0,clip]{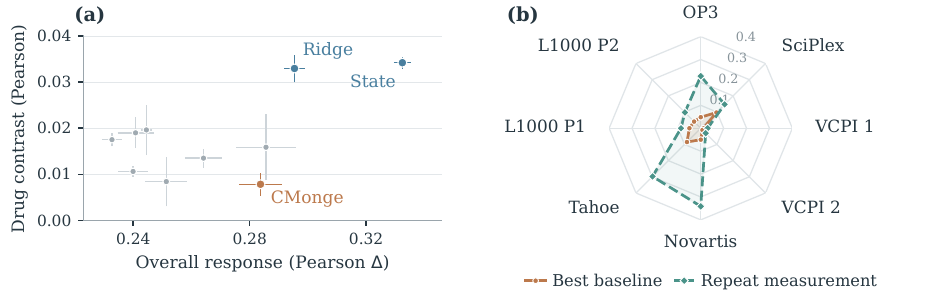}
\vspace{-6pt}
\caption{\textbf{Overall agreement and drug-specific prediction.} \textbf{(a)} Overall response and drug contrast for ten complete-coverage baselines: three-split means and sample SDs over 14 native views. \textbf{(b)} Drug contrast of the best available baseline and a same-drug repeat, averaged over views and splits on matched repeat-eligible samples. Novartis uses HVG; VCPI repeats are within-plate wells.}
\label{fig:benchmark_findings}
\vspace{0pt}
\end{figure}

\noindent\textbf{Overall agreement does not imply drug-specific prediction.}
Strong overall-response agreement does not ensure accurate prediction of drug differences: CMonge ranks fourth on overall response but last on drug contrast (Figure~\ref{fig:benchmark_findings}a).
To interpret these low contrast scores, we compare model performance with the agreement between repeated measurements.
Even the strongest baseline in each dataset falls below same-drug repeat agreement on matched samples (Figure~\ref{fig:benchmark_findings}b), motivating direct supervision of these reproducible drug differences.

\noindent\textbf{Chemical supervision is fragmented across screens.}
Complementary screens offer additional molecular supervision for learning these differences.
However, 89.6\% of benchmark compounds occur in only one dataset, so training on individual screens leaves this complementary chemistry unused.
Pooling TRAIN compounds across datasets increases the fraction of validation compounds with a close structural analog (Tanimoto $\geq 0.5$) from 7\% to 47\% on OP3 and from 24\% to 71\% on Tahoe.
Direct cross-dataset neighbor copying nevertheless lowers mean contrast from 0.046 to 0.040 (Appendix~\ref{app:neighbor_copying}), motivating learned mappings.

\section{Methods}
\label{sec:methods}
\begingroup
\setlength{\parskip}{3pt}
\newenvironment{methodequation}{%
  \begingroup\small
  \setlength{\abovedisplayskip}{3pt plus 1pt minus 1pt}%
  \setlength{\belowdisplayskip}{3pt plus 1pt minus 1pt}%
  \setlength{\abovedisplayshortskip}{3pt plus 1pt}%
  \setlength{\belowdisplayshortskip}{3pt plus 1pt}%
  \setlength{\jot}{2pt}%
  \begin{equation}}{\end{equation}\endgroup\ignorespacesafterend}

Bison connects native gene panels through a shared representation, enabling joint learning from chemical supervision across datasets.
Two discrete diffusion models (DLMs) learn complete responses and explicitly supervised molecular differences, respectively; their predictions are composed as a context-dependent reference plus a molecular deviation (Figure~\ref{fig:architecture}).

\begin{figure}[!t]
    \centering
    \vspace{-5pt}
    \includegraphics[width=\textwidth,trim=0 14bp 0 8bp,clip]{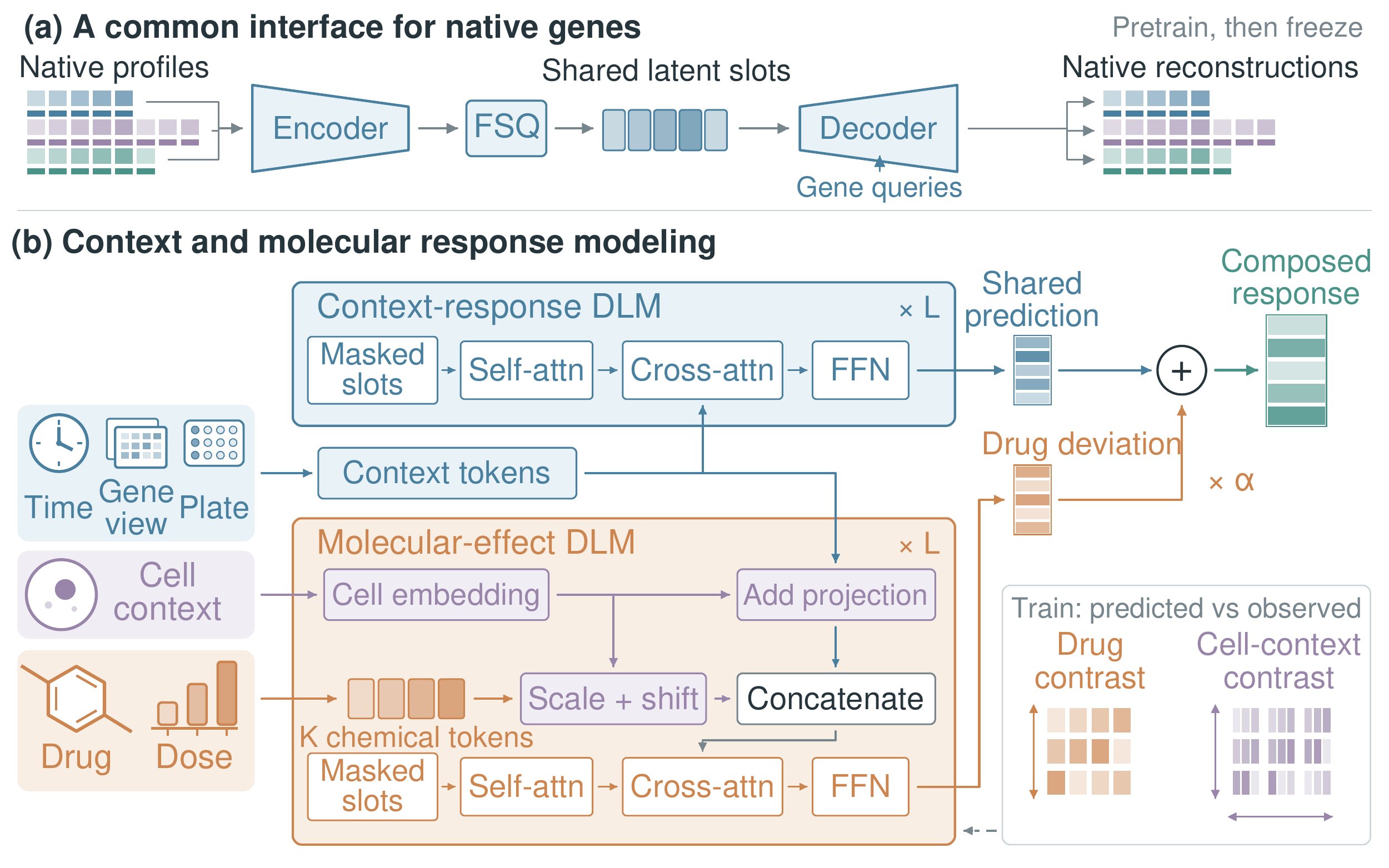}
    \vspace{-5pt}
    \caption{\textbf{Bison architecture.} (a) Shared encoding and gene-query decoding connect native panels. (b) The molecular DLM uses cellular context to modulate chemical tokens and condition context memory and learns both within-context drug differences and drug--context interactions. The shared prediction includes the control reference; the drug deviation subtracts a control-coded molecular prediction. Their sum is calibrated afterward. $L$ and $K$ denote layer and chemical-token counts; training losses are defined in Section~\ref{sec:contrast_learning}.}
    \label{fig:architecture}
    \vspace{-3pt}
\end{figure}

\subsection{Shared Encoding and Decoding}
\label{sec:gene_interface}

Joint learning must connect different gene panels without discarding measurements outside their intersection. We represent a profile as a set of gene identities and values. An attention-based encoder pools this set~\citep{lee2019set}, finite scalar quantization (FSQ) creates fixed-length discrete codes~\citep{mentzer2024fsq}, and gene-identity queries decode the requested native panel:
\begin{methodequation}
\mathbf z=Q\!\left(E_\phi\!\left(\{(g,s_g):g\in G\}\right)\right),
\qquad \hat{\mathbf s}_G=D_\psi(\mathbf z,G).
\label{eq:gene_interface}
\end{methodequation}
Here $G$ is the native gene set, $c$ collects the experimental conditions, and $\mathbf s_G=a_d(\mathbf x_G-\boldsymbol\mu_{c,G})$ is a scaled expression residual. The fixed dataset scale $a_d$ accounts for expression-scale differences. The empirical-Bayes reference $\boldsymbol\mu_{c,G}$ uses TRAIN controls, shrinking the plate mean toward the cellular-context mean according to control precision to reduce fluctuations from small control sets. We pretrain the encoder and decoder by joint reconstruction on Full and HVG panels, then freeze them.
Both DLMs share stable coordinates while retaining each view's input and output genes.

\subsection{Modeling the Shared Response}
\label{sec:conditional_model}

Drug-specific changes occur on top of cellular and experimental structure that a complete prediction must retain. The context-response DLM learns this structure from complete conditional profiles. Matched-control features convey cellular background, plate offsets and control reliability; time and native-view information specify the remaining context. Following masked diffusion~\citep{sahoo2024masked}, categorical cross-entropy on masked expression codes, $\mathcal L_{\mathrm{tok}}$, trains the model to reconstruct profiles, including fully masked examples that require prediction from conditions alone. A decoded-response loss, $\mathcal L_{\mathrm{resp}}$, additionally minimizes $1-\rho$ between predicted and observed control-matched condition means, where $\rho$ is correlation across genes. This ties code prediction to the expression response measured by the benchmark.

At composition time, we fix this model's molecular input to a reference condition. Its output then varies with experimental context while remaining shared across query compounds.
This retains the learned response structure and assigns compound-dependent effects to the second DLM.

\subsection{Learning Molecular Differences}
\label{sec:contrast_learning}

The same molecule can elicit different responses in different cell types. The molecular-effect DLM therefore combines molecular structure with explicit cellular context: Morgan fingerprints~\citep{rogers2010ecfp}, dose and time form chemical attention tokens, whose scale and shift depend on a learned cell-context embedding. The embedding also contributes to the model's context memory. This allows chemical conditioning to change with cellular background; two additional losses require the resulting predictions to preserve the relevant differences.

\paragraph{Distinguishing drugs within a context.} We compare compounds under the same cellular background, plate, dose and time, averaging replicates for each compound. For $K$ sampled compounds, we center the observed and predicted control-matched responses separately:
\begin{methodequation}
\mathbf d_i=\boldsymbol\delta_i-\bar{\boldsymbol\delta},\qquad
\hat{\mathbf d}_i=\hat{\boldsymbol\delta}_i-\bar{\hat{\boldsymbol\delta}},\qquad
\mathcal L_{\mathrm{ctr}}=\frac1K\sum_{i=1}^{K}\!\left[1-\rho(\hat{\mathbf d}_i,\mathbf d_i)\right].
\label{eq:contrast_loss}
\end{methodequation}
Bars denote means across sampled compounds. Centering removes the shared additive response, so matching background alone cannot reduce this loss.
Comparisons use native panels, enabling joint supervision without matched compounds or identical genes across studies.

\paragraph{Distinguishing context-dependent drug effects.} A model could still assign each drug the same effect across cell types. To supervise what changes with cellular context, we also sample matched drug-by-context grids under the same plate, dose and time. For $K_d$ drugs and $K_c$ contexts, we remove both marginal means:
\begin{methodequation}
\mathbf v_{i\ell}=\boldsymbol\delta_{i\ell}-\bar{\boldsymbol\delta}_{i\cdot}-\bar{\boldsymbol\delta}_{\cdot\ell}+\bar{\boldsymbol\delta}_{\cdot\cdot},\qquad
\mathcal L_{\mathrm{cell}}=\frac1{K_dK_c}\sum_{i,\ell}\!\left[1-\rho(\hat{\mathbf v}_{i\ell},\mathbf v_{i\ell})\right].
\label{eq:cell_context_loss}
\end{methodequation}
Dots indicate averaged indices, and predictions are centered identically. The remaining interaction excludes both a drug effect common to all contexts and a context effect common to all drugs. This loss applies only where matched grids exist, here OP3 and Tahoe; it does not assume measurements of every drug in every dataset.

The context and molecular DLMs, with separate parameters $\eta$ and $\theta$, minimize
\begin{methodequation}
\mathcal L_\eta=\mathcal L_{\mathrm{tok}}+\lambda_{\mathrm{resp}}\sum_u\mathcal L_{\mathrm{resp}}^{(u)},\qquad
\mathcal L_\theta=\mathcal L_{\mathrm{tok}}+\sum_u\!\left(\lambda_{\mathrm{resp}}\mathcal L_{\mathrm{resp}}^{(u)}+\lambda_{\mathrm{ctr}}\mathcal L_{\mathrm{ctr}}^{(u)}+\lambda_{\mathrm{cell}}\mathcal L_{\mathrm{cell}}^{(u)}\right),
\label{eq:training_objective}
\end{methodequation}
where $u$ indexes native training views, $\lambda_{\mathrm{resp}}=2.5$, and $\lambda_{\mathrm{ctr}}=\lambda_{\mathrm{cell}}=10$.
The interaction term is omitted for datasets without eligible grids. All decoded losses use fully masked inputs, expression-scale restoration, nonnegative clipping and matched TRAIN controls before replicate aggregation; gradients pass through the frozen decoder. The two models are selected on VAL overall-response and drug-contrast correlation, respectively. Sampling, training and decoding details are in Appendix~\ref{app:bison_training}.

\subsection{Composing the Predictions}
\label{sec:composition}

The shared decoder returns both models' outputs to the requested genes and expression scale, making them directly composable. One-step soft decoding predicts each FSQ digit's distribution from a fully masked input and decodes its expected coordinate into an expression residual. Write these residuals as $\mathbf r_\eta$ and $\mathbf r_\theta$ for the context and molecular DLMs.

The molecular DLM is evaluated at both the query drug $m$ and a dataset-specific fixed reference $m_0$, retaining the same cell-context embedding. The reference conditions $c_0$ preserve matched controls, time and native view, with dose and treatment status set to control values.
Subtracting this reference isolates the molecular deviation and avoids duplicating the reference response:
\begin{methodequation}
\tilde{\mathbf x}_G=\boldsymbol\mu_{c,G}
+\mathbf r_\eta(m_0,c,G)
+\alpha\!\left[\mathbf r_\theta(m,c,G)-\mathbf r_\theta(m_0,c_0,G)\right].
\label{eq:composed_prediction}
\end{methodequation}
The globally fixed $\alpha=0.25$ controls the contribution of molecular differences. We restore the reference once and retain the untruncated composition for calibration. A prediction uses one context-model call and two molecular-model calls, all with the same frozen decoder.

\subsection{Response Calibration}
\label{sec:postprocessing}

Correct response direction does not ensure correct magnitude.
For each count-based dataset and view, we estimate a scale from TRAIN condition-mean effects $E$ and $\hat E$ relative to matched plate controls, balancing absolute-magnitude matching and least-squares fit:
\begin{methodequation}
s_{\mathrm{abs}}=\frac{\|E\|_1}{\|\hat E\|_1},\qquad
s_{\mathrm{LS}}=\frac{\langle E,\hat E\rangle_F}{\|\hat E\|_F^2},\qquad
s=\sqrt{s_{\mathrm{abs}}\max(s_{\mathrm{LS}},\epsilon)}.
\label{eq:response_calibration}
\end{methodequation}
With matched control mean $\mathbf b_i$, the calibrated prediction is $\mathbf x_i^{\mathrm{cal}}=[\mathbf b_i+s(\tilde{\mathbf x}_i-\mathbf b_i)]_+$, scaling the complete response before nonnegative clipping.

Continuous decoding can also leave small positive values for rarely detected genes.
We set a prediction to zero when the gene's TRAIN detection frequency is below 10\% and the value falls below half the median per-profile minimum positive TRAIN expression.
This retains weak responses in frequently detected genes.
The rules are shared across count-based Full/HVG panels, with quantities estimated from TRAIN (Appendix~\ref{app:calibration_details}); L1000 retains the uncalibrated composition with nonnegative clipping.

\endgroup

\endgroup
\section{Experiments}\label{sec:experiments}
We evaluate whether joint learning improves unseen-compound prediction, where cross-dataset supervision helps, and which model components contribute, following the protocol in Section~\ref{sec:benchmark}.

\subsection{Comparison with Existing Perturbation Models}\label{sec:overall_performance}

We compare one jointly trained Bison model with 11 dataset-specific methods and two controls, Linear and Global Mean, across eight datasets and fourteen native views.
Table~\ref{tab:benchmark_results} reports three-split means and sample SD; Appendix~\ref{app:multiseed_benchmark} gives per-view results.

\begin{table}[H]
\centering
\caption{\textbf{Benchmark results on globally unseen compounds.}
Pearson $\times100$, mean (sample SD) over three compound splits.
Dataset columns average Full/HVG (L1000: landmark genes); Macro weights fourteen native views equally.
Top: plate-matched overall response; bottom: drug contrast.
Bold/underline: best/second-best complete-coverage mean.
$^{*}$Partial coverage, not ranked; dashes: unavailable; NA: undefined.
Per-view results and resource limits are in Appendices~\ref{app:baseline_adaptations}--\ref{app:experiment_details}.}
\label{tab:benchmark_results}
\footnotesize
\setlength{\tabcolsep}{0pt}
\begin{tabular*}{\textwidth}{@{\extracolsep{\fill}}lrrrrrrrrr}
\toprule
Method & OP3 & SciPlex & VCPI1 & VCPI2 & Novartis & Tahoe & P1 & P2 & Macro \\
\midrule
\multicolumn{10}{l}{\textit{Overall response}} \\
Global Mean & 7.1{\fontsize{6}{7}\selectfont\color{black!55}\,(1.1)} & 17.8{\fontsize{6}{7}\selectfont\color{black!55}\,(0.7)} & 18.1{\fontsize{6}{7}\selectfont\color{black!55}\,(0.3)} & 15.5{\fontsize{6}{7}\selectfont\color{black!55}\,(0.1)} & 29.8{\fontsize{6}{7}\selectfont\color{black!55}\,(0.8)} & 20.3{\fontsize{6}{7}\selectfont\color{black!55}\,(1.1)} & 8.7{\fontsize{6}{7}\selectfont\color{black!55}\,(0.0)} & 4.1{\fontsize{6}{7}\selectfont\color{black!55}\,(0.1)} & 16.4{\fontsize{6}{7}\selectfont\color{black!55}\,(0.2)} \\
Linear & 28.9{\fontsize{6}{7}\selectfont\color{black!55}\,(2.9)} & 44.6{\fontsize{6}{7}\selectfont\color{black!55}\,(1.4)} & 12.7{\fontsize{6}{7}\selectfont\color{black!55}\,(0.1)} & 5.9{\fontsize{6}{7}\selectfont\color{black!55}\,(0.5)} & 25.4{\fontsize{6}{7}\selectfont\color{black!55}\,(0.9)} & 36.2{\fontsize{6}{7}\selectfont\color{black!55}\,(2.7)} & 15.6{\fontsize{6}{7}\selectfont\color{black!55}\,(0.3)} & 2.9{\fontsize{6}{7}\selectfont\color{black!55}\,(0.3)} & 23.3{\fontsize{6}{7}\selectfont\color{black!55}\,(0.3)} \\
Ridge & \underline{39.0}{\fontsize{6}{7}\selectfont\color{black!55}\,(1.6)} & 49.8{\fontsize{6}{7}\selectfont\color{black!55}\,(0.6)} & 18.5{\fontsize{6}{7}\selectfont\color{black!55}\,(0.4)} & 15.7{\fontsize{6}{7}\selectfont\color{black!55}\,(0.2)} & 30.7{\fontsize{6}{7}\selectfont\color{black!55}\,(0.8)} & 42.0{\fontsize{6}{7}\selectfont\color{black!55}\,(1.0)} & 15.6{\fontsize{6}{7}\selectfont\color{black!55}\,(0.3)} & 6.3{\fontsize{6}{7}\selectfont\color{black!55}\,(0.1)} & 29.5{\fontsize{6}{7}\selectfont\color{black!55}\,(0.3)} \\
chemCPA & 33.1{\fontsize{6}{7}\selectfont\color{black!55}\,(0.7)} & 44.1{\fontsize{6}{7}\selectfont\color{black!55}\,(0.7)} & 12.7{\fontsize{6}{7}\selectfont\color{black!55}\,(0.1)} & 11.0{\fontsize{6}{7}\selectfont\color{black!55}\,(0.2)} & 23.4{\fontsize{6}{7}\selectfont\color{black!55}\,(1.6)} & 36.8{\fontsize{6}{7}\selectfont\color{black!55}\,(1.7)} & 14.3{\fontsize{6}{7}\selectfont\color{black!55}\,(0.2)} & 6.0{\fontsize{6}{7}\selectfont\color{black!55}\,(0.6)} & 24.5{\fontsize{6}{7}\selectfont\color{black!55}\,(0.2)} \\
PRnet & 26.4{\fontsize{6}{7}\selectfont\color{black!55}\,(1.5)} & 44.3{\fontsize{6}{7}\selectfont\color{black!55}\,(0.6)} & 12.5{\fontsize{6}{7}\selectfont\color{black!55}\,(3.2)} & 13.4{\fontsize{6}{7}\selectfont\color{black!55}\,(0.1)} & 28.5{\fontsize{6}{7}\selectfont\color{black!55}\,(0.3)} & 33.1{\fontsize{6}{7}\selectfont\color{black!55}\,(1.5)} & 15.5{\fontsize{6}{7}\selectfont\color{black!55}\,(0.2)} & 5.1{\fontsize{6}{7}\selectfont\color{black!55}\,(0.2)} & 24.1{\fontsize{6}{7}\selectfont\color{black!55}\,(0.6)} \\
biolord & 34.4{\fontsize{6}{7}\selectfont\color{black!55}\,(6.1)} & 48.8{\fontsize{6}{7}\selectfont\color{black!55}\,(2.1)} & 17.6{\fontsize{6}{7}\selectfont\color{black!55}\,(0.1)} & 14.9{\fontsize{6}{7}\selectfont\color{black!55}\,(0.4)} & 29.1{\fontsize{6}{7}\selectfont\color{black!55}\,(0.8)} & 42.5{\fontsize{6}{7}\selectfont\color{black!55}\,(1.0)} & 19.2{\fontsize{6}{7}\selectfont\color{black!55}\,(0.8)} & 6.2{\fontsize{6}{7}\selectfont\color{black!55}\,(0.3)} & 28.6{\fontsize{6}{7}\selectfont\color{black!55}\,(1.0)} \\
PerturbNet & 32.2{\fontsize{6}{7}\selectfont\color{black!55}\,(2.1)} & 46.1{\fontsize{6}{7}\selectfont\color{black!55}\,(2.0)} & 17.5{\fontsize{6}{7}\selectfont\color{black!55}\,(0.6)} & 15.1{\fontsize{6}{7}\selectfont\color{black!55}\,(0.5)} & 28.0{\fontsize{6}{7}\selectfont\color{black!55}\,(0.9)} & 20.9{\fontsize{6}{7}\selectfont\color{black!55}\,(2.2)} & 11.4{\fontsize{6}{7}\selectfont\color{black!55}\,(4.2)} & 4.9{\fontsize{6}{7}\selectfont\color{black!55}\,(0.2)} & 24.0{\fontsize{6}{7}\selectfont\color{black!55}\,(0.5)} \\
CellFlow & 38.8{\fontsize{6}{7}\selectfont\color{black!55}\,(5.3)} & \underline{50.4}{\fontsize{6}{7}\selectfont\color{black!55}\,(0.5)} & 18.6{\fontsize{6}{7}\selectfont\color{black!55}\,(0.4)} & 15.9{\fontsize{6}{7}\selectfont\color{black!55}\,(0.4)} & 29.3{\fontsize{6}{7}\selectfont\color{black!55}\,(0.5)} & 44.3{\fontsize{6}{7}\selectfont\color{black!55}\,(1.5)} & -- & -- & \textcolor{black!55}{32.9$^{*}$}{\fontsize{6}{7}\selectfont\color{black!55}\,(1.1)} \\
CMonge & 34.8{\fontsize{6}{7}\selectfont\color{black!55}\,(4.8)} & \textbf{50.7}{\fontsize{6}{7}\selectfont\color{black!55}\,(0.3)} & 18.0{\fontsize{6}{7}\selectfont\color{black!55}\,(0.3)} & 15.5{\fontsize{6}{7}\selectfont\color{black!55}\,(0.2)} & 29.1{\fontsize{6}{7}\selectfont\color{black!55}\,(0.8)} & 41.2{\fontsize{6}{7}\selectfont\color{black!55}\,(0.5)} & 13.0{\fontsize{6}{7}\selectfont\color{black!55}\,(0.3)} & 5.7{\fontsize{6}{7}\selectfont\color{black!55}\,(0.2)} & 28.4{\fontsize{6}{7}\selectfont\color{black!55}\,(0.7)} \\
Prophet & 37.1{\fontsize{6}{7}\selectfont\color{black!55}\,(2.0)} & 48.0{\fontsize{6}{7}\selectfont\color{black!55}\,(0.9)} & 16.9{\fontsize{6}{7}\selectfont\color{black!55}\,(0.2)} & 15.3{\fontsize{6}{7}\selectfont\color{black!55}\,(0.4)} & 28.0{\fontsize{6}{7}\selectfont\color{black!55}\,(1.0)} & 29.9{\fontsize{6}{7}\selectfont\color{black!55}\,(1.3)} & 14.1{\fontsize{6}{7}\selectfont\color{black!55}\,(0.1)} & 5.5{\fontsize{6}{7}\selectfont\color{black!55}\,(0.3)} & 26.4{\fontsize{6}{7}\selectfont\color{black!55}\,(0.6)} \\
PrePR-CT & 36.9{\fontsize{6}{7}\selectfont\color{black!55}\,(0.4)} & 37.5{\fontsize{6}{7}\selectfont\color{black!55}\,(4.2)} & 17.6{\fontsize{6}{7}\selectfont\color{black!55}\,(0.3)} & 15.1{\fontsize{6}{7}\selectfont\color{black!55}\,(0.2)} & 30.2{\fontsize{6}{7}\selectfont\color{black!55}\,(1.0)} & 32.1{\fontsize{6}{7}\selectfont\color{black!55}\,(1.3)} & 9.0{\fontsize{6}{7}\selectfont\color{black!55}\,(0.6)} & 4.3{\fontsize{6}{7}\selectfont\color{black!55}\,(0.3)} & 25.1{\fontsize{6}{7}\selectfont\color{black!55}\,(0.7)} \\
XPert & 39.0{\fontsize{6}{7}\selectfont\color{black!55}\,(1.5)} & \textcolor{black!55}{46.3$^{*}$}{\fontsize{6}{7}\selectfont\color{black!55}\,(3.4)} & \textcolor{black!55}{20.1$^{*}$}{\fontsize{6}{7}\selectfont\color{black!55}\,(20.3)} & \textcolor{black!55}{21.5$^{*}$}{\fontsize{6}{7}\selectfont\color{black!55}\,(19.9)} & \textcolor{black!55}{56.5$^{*}$}{\fontsize{6}{7}\selectfont\color{black!55}\,(0.7)} & \textcolor{black!55}{50.6$^{*}$}{\fontsize{6}{7}\selectfont\color{black!55}\,(1.4)} & \underline{29.6}{\fontsize{6}{7}\selectfont\color{black!55}\,(0.2)} & \underline{15.6}{\fontsize{6}{7}\selectfont\color{black!55}\,(1.8)} & \textcolor{black!55}{35.4$^{*}$}{\fontsize{6}{7}\selectfont\color{black!55}\,(0.3)} \\
State & 29.3{\fontsize{6}{7}\selectfont\color{black!55}\,(2.9)} & 39.8{\fontsize{6}{7}\selectfont\color{black!55}\,(0.8)} & \underline{23.5}{\fontsize{6}{7}\selectfont\color{black!55}\,(0.7)} & \underline{23.7}{\fontsize{6}{7}\selectfont\color{black!55}\,(1.0)} & \underline{48.2}{\fontsize{6}{7}\selectfont\color{black!55}\,(0.5)} & \underline{48.9}{\fontsize{6}{7}\selectfont\color{black!55}\,(1.1)} & 27.2{\fontsize{6}{7}\selectfont\color{black!55}\,(0.2)} & 11.7{\fontsize{6}{7}\selectfont\color{black!55}\,(1.4)} & \underline{33.2}{\fontsize{6}{7}\selectfont\color{black!55}\,(0.3)} \\
\midrule \textbf{Bison} & \textbf{42.1}{\fontsize{6}{7}\selectfont\color{black!55}\,(2.4)} & 46.1{\fontsize{6}{7}\selectfont\color{black!55}\,(1.3)} & \textbf{32.6}{\fontsize{6}{7}\selectfont\color{black!55}\,(0.9)} & \textbf{32.9}{\fontsize{6}{7}\selectfont\color{black!55}\,(0.6)} & \textbf{51.6}{\fontsize{6}{7}\selectfont\color{black!55}\,(1.0)} & \textbf{55.0}{\fontsize{6}{7}\selectfont\color{black!55}\,(0.4)} & \textbf{34.3}{\fontsize{6}{7}\selectfont\color{black!55}\,(1.0)} & \textbf{20.7}{\fontsize{6}{7}\selectfont\color{black!55}\,(0.7)} & \textbf{41.1}{\fontsize{6}{7}\selectfont\color{black!55}\,(0.6)} \\
\midrule
\multicolumn{10}{l}{\textit{Drug contrast}} \\
Global Mean & NA & NA & NA & NA & NA & NA & NA & NA & NA \\
Linear & 0.9{\fontsize{6}{7}\selectfont\color{black!55}\,(1.5)} & 5.5{\fontsize{6}{7}\selectfont\color{black!55}\,(1.0)} & 0.3{\fontsize{6}{7}\selectfont\color{black!55}\,(0.3)} & 0.1{\fontsize{6}{7}\selectfont\color{black!55}\,(0.5)} & 2.0{\fontsize{6}{7}\selectfont\color{black!55}\,(0.3)} & 2.2{\fontsize{6}{7}\selectfont\color{black!55}\,(1.7)} & 2.0{\fontsize{6}{7}\selectfont\color{black!55}\,(0.4)} & 0.5{\fontsize{6}{7}\selectfont\color{black!55}\,(0.4)} & 1.8{\fontsize{6}{7}\selectfont\color{black!55}\,(0.1)} \\
Ridge & 4.2{\fontsize{6}{7}\selectfont\color{black!55}\,(4.4)} & 7.7{\fontsize{6}{7}\selectfont\color{black!55}\,(1.9)} & \underline{1.5}{\fontsize{6}{7}\selectfont\color{black!55}\,(0.5)} & \underline{1.4}{\fontsize{6}{7}\selectfont\color{black!55}\,(0.3)} & \textbf{4.0}{\fontsize{6}{7}\selectfont\color{black!55}\,(0.4)} & 2.1{\fontsize{6}{7}\selectfont\color{black!55}\,(0.9)} & 2.2{\fontsize{6}{7}\selectfont\color{black!55}\,(0.4)} & 1.9{\fontsize{6}{7}\selectfont\color{black!55}\,(0.2)} & 3.3{\fontsize{6}{7}\selectfont\color{black!55}\,(0.3)} \\
chemCPA & 0.9{\fontsize{6}{7}\selectfont\color{black!55}\,(2.3)} & \underline{8.8}{\fontsize{6}{7}\selectfont\color{black!55}\,(2.3)} & 0.8{\fontsize{6}{7}\selectfont\color{black!55}\,(0.5)} & 0.2{\fontsize{6}{7}\selectfont\color{black!55}\,(0.2)} & 0.5{\fontsize{6}{7}\selectfont\color{black!55}\,(0.1)} & 1.0{\fontsize{6}{7}\selectfont\color{black!55}\,(2.5)} & 2.0{\fontsize{6}{7}\selectfont\color{black!55}\,(0.8)} & 1.2{\fontsize{6}{7}\selectfont\color{black!55}\,(0.5)} & 2.0{\fontsize{6}{7}\selectfont\color{black!55}\,(0.5)} \\
PRnet & 0.0{\fontsize{6}{7}\selectfont\color{black!55}\,(2.3)} & \textbf{9.7}{\fontsize{6}{7}\selectfont\color{black!55}\,(2.9)} & 0.3{\fontsize{6}{7}\selectfont\color{black!55}\,(0.1)} & 0.4{\fontsize{6}{7}\selectfont\color{black!55}\,(0.2)} & 1.3{\fontsize{6}{7}\selectfont\color{black!55}\,(0.1)} & 0.9{\fontsize{6}{7}\selectfont\color{black!55}\,(0.7)} & 1.2{\fontsize{6}{7}\selectfont\color{black!55}\,(0.3)} & 0.3{\fontsize{6}{7}\selectfont\color{black!55}\,(0.2)} & 1.9{\fontsize{6}{7}\selectfont\color{black!55}\,(0.3)} \\
biolord & -0.3{\fontsize{6}{7}\selectfont\color{black!55}\,(1.0)} & 7.1{\fontsize{6}{7}\selectfont\color{black!55}\,(5.0)} & 0.3{\fontsize{6}{7}\selectfont\color{black!55}\,(0.3)} & -0.0{\fontsize{6}{7}\selectfont\color{black!55}\,(0.3)} & 1.0{\fontsize{6}{7}\selectfont\color{black!55}\,(0.2)} & 1.9{\fontsize{6}{7}\selectfont\color{black!55}\,(1.3)} & 1.4{\fontsize{6}{7}\selectfont\color{black!55}\,(0.4)} & 0.8{\fontsize{6}{7}\selectfont\color{black!55}\,(0.4)} & 1.6{\fontsize{6}{7}\selectfont\color{black!55}\,(0.7)} \\
PerturbNet & 0.4{\fontsize{6}{7}\selectfont\color{black!55}\,(1.3)} & 3.6{\fontsize{6}{7}\selectfont\color{black!55}\,(2.0)} & 0.5{\fontsize{6}{7}\selectfont\color{black!55}\,(0.4)} & 0.5{\fontsize{6}{7}\selectfont\color{black!55}\,(0.3)} & 1.7{\fontsize{6}{7}\selectfont\color{black!55}\,(0.3)} & 0.3{\fontsize{6}{7}\selectfont\color{black!55}\,(0.4)} & 0.2{\fontsize{6}{7}\selectfont\color{black!55}\,(0.3)} & 0.5{\fontsize{6}{7}\selectfont\color{black!55}\,(0.2)} & 1.1{\fontsize{6}{7}\selectfont\color{black!55}\,(0.1)} \\
CellFlow & -1.8{\fontsize{6}{7}\selectfont\color{black!55}\,(1.3)} & 6.7{\fontsize{6}{7}\selectfont\color{black!55}\,(2.0)} & 0.6{\fontsize{6}{7}\selectfont\color{black!55}\,(0.5)} & 0.4{\fontsize{6}{7}\selectfont\color{black!55}\,(0.3)} & 1.4{\fontsize{6}{7}\selectfont\color{black!55}\,(0.7)} & 1.0{\fontsize{6}{7}\selectfont\color{black!55}\,(1.2)} & -- & -- & \textcolor{black!55}{1.4$^{*}$}{\fontsize{6}{7}\selectfont\color{black!55}\,(0.1)} \\
CMonge & -0.6{\fontsize{6}{7}\selectfont\color{black!55}\,(1.0)} & 4.9{\fontsize{6}{7}\selectfont\color{black!55}\,(2.1)} & -0.0{\fontsize{6}{7}\selectfont\color{black!55}\,(0.3)} & 0.2{\fontsize{6}{7}\selectfont\color{black!55}\,(0.2)} & 0.6{\fontsize{6}{7}\selectfont\color{black!55}\,(0.2)} & 0.4{\fontsize{6}{7}\selectfont\color{black!55}\,(0.3)} & 0.1{\fontsize{6}{7}\selectfont\color{black!55}\,(0.1)} & 0.3{\fontsize{6}{7}\selectfont\color{black!55}\,(0.4)} & 0.8{\fontsize{6}{7}\selectfont\color{black!55}\,(0.3)} \\
Prophet & 0.6{\fontsize{6}{7}\selectfont\color{black!55}\,(1.8)} & 4.1{\fontsize{6}{7}\selectfont\color{black!55}\,(1.0)} & 0.7{\fontsize{6}{7}\selectfont\color{black!55}\,(0.4)} & 0.3{\fontsize{6}{7}\selectfont\color{black!55}\,(0.3)} & 0.8{\fontsize{6}{7}\selectfont\color{black!55}\,(0.3)} & 2.2{\fontsize{6}{7}\selectfont\color{black!55}\,(1.1)} & 0.7{\fontsize{6}{7}\selectfont\color{black!55}\,(0.2)} & 0.9{\fontsize{6}{7}\selectfont\color{black!55}\,(0.4)} & 1.4{\fontsize{6}{7}\selectfont\color{black!55}\,(0.2)} \\
PrePR-CT & -0.3{\fontsize{6}{7}\selectfont\color{black!55}\,(2.3)} & 5.0{\fontsize{6}{7}\selectfont\color{black!55}\,(1.9)} & -0.1{\fontsize{6}{7}\selectfont\color{black!55}\,(0.5)} & 0.1{\fontsize{6}{7}\selectfont\color{black!55}\,(0.2)} & 0.8{\fontsize{6}{7}\selectfont\color{black!55}\,(0.2)} & 0.4{\fontsize{6}{7}\selectfont\color{black!55}\,(0.1)} & -0.1{\fontsize{6}{7}\selectfont\color{black!55}\,(0.1)} & 0.3{\fontsize{6}{7}\selectfont\color{black!55}\,(0.4)} & 0.8{\fontsize{6}{7}\selectfont\color{black!55}\,(0.5)} \\
XPert & -0.6{\fontsize{6}{7}\selectfont\color{black!55}\,(0.5)} & \textcolor{black!55}{4.7$^{*}$}{\fontsize{6}{7}\selectfont\color{black!55}\,(1.1)} & -- & \textcolor{black!55}{0.9$^{*}$}{\fontsize{6}{7}\selectfont\color{black!55}\,(1.1)} & \textcolor{black!55}{5.0$^{*}$}{\fontsize{6}{7}\selectfont\color{black!55}\,(0.7)} & \textcolor{black!55}{1.7$^{*}$}{\fontsize{6}{7}\selectfont\color{black!55}\,(1.3)} & \underline{4.9}{\fontsize{6}{7}\selectfont\color{black!55}\,(1.0)} & \underline{4.1}{\fontsize{6}{7}\selectfont\color{black!55}\,(0.2)} & \textcolor{black!55}{2.5$^{*}$}{\fontsize{6}{7}\selectfont\color{black!55}\,(0.3)} \\
State & \underline{4.8}{\fontsize{6}{7}\selectfont\color{black!55}\,(1.9)} & 6.2{\fontsize{6}{7}\selectfont\color{black!55}\,(1.3)} & 1.4{\fontsize{6}{7}\selectfont\color{black!55}\,(0.2)} & 1.4{\fontsize{6}{7}\selectfont\color{black!55}\,(0.3)} & 3.3{\fontsize{6}{7}\selectfont\color{black!55}\,(0.3)} & \textbf{3.8}{\fontsize{6}{7}\selectfont\color{black!55}\,(0.8)} & 3.3{\fontsize{6}{7}\selectfont\color{black!55}\,(0.6)} & 2.7{\fontsize{6}{7}\selectfont\color{black!55}\,(0.1)} & \underline{3.4}{\fontsize{6}{7}\selectfont\color{black!55}\,(0.1)} \\
\midrule \textbf{Bison} & \textbf{9.1}{\fontsize{6}{7}\selectfont\color{black!55}\,(1.9)} & 7.8{\fontsize{6}{7}\selectfont\color{black!55}\,(2.5)} & \textbf{3.0}{\fontsize{6}{7}\selectfont\color{black!55}\,(0.3)} & \textbf{2.0}{\fontsize{6}{7}\selectfont\color{black!55}\,(0.6)} & \underline{3.9}{\fontsize{6}{7}\selectfont\color{black!55}\,(1.1)} & \underline{3.0}{\fontsize{6}{7}\selectfont\color{black!55}\,(2.3)} & \textbf{5.0}{\fontsize{6}{7}\selectfont\color{black!55}\,(1.1)} & \textbf{4.2}{\fontsize{6}{7}\selectfont\color{black!55}\,(0.7)} & \textbf{4.8}{\fontsize{6}{7}\selectfont\color{black!55}\,(0.4)} \\
\bottomrule
\end{tabular*}
\end{table}

One jointly trained Bison model achieves the highest complete-coverage means on both metrics.
Overall-response Pearson reaches 0.4110 versus 0.3325 for State, with Bison leading on eleven views.
The gain extends to drug-specific differences: drug contrast reaches 0.0478 versus 0.0342, with eight view-level wins.
The dataset-level comparison is more varied: competing methods achieve higher drug contrast on SciPlex, Novartis and Tahoe.

The official Cell-Eval metrics assess complementary aspects of the final prediction (Table~\ref{tab:official_metrics}).
Bison achieves the highest mean Pearson $\Delta$ and perturbation-discrimination scores and the lowest MAE across twelve count-based views.
Differential-expression recovery remains mixed: State retains higher overlap@100, and Prophet has higher direction agreement.

\begin{table}[!htbp]
\centering
\caption{\textbf{Official Cell-Eval results.} Mean (sample SD) across three compound splits; values $\times100$. Bison includes amplitude calibration and detection-aware zeroing. Twelve Full/HVG views of six count-based datasets; DE direction uses ten views. Bold: best complete-view mean. $^*$Partial view coverage, unranked. $^\dagger$Some evaluations use valid-condition subsets; Bison uses all conditions.}
\label{tab:official_metrics}
\label{tab:cell_eval_multiseed}
\begingroup
\definecolor{Muted}{HTML}{78818A}
\fontsize{8}{9.6}\selectfont
\setlength{\tabcolsep}{2pt}
\renewcommand{\arraystretch}{1.13}
\begin{tabular*}{\textwidth}{@{\extracolsep{\fill}}lc*{7}{r}@{}}
\toprule
Method & Views & \shortstack{Pearson\\$\Delta\uparrow$} & \shortstack{MAE\\$\Delta\downarrow$} & \shortstack{PDS\\L1 $\uparrow$} & \shortstack{PDS\\L2 $\uparrow$} & \shortstack{PDS\\cosine $\uparrow$} & \shortstack{DE\\direction $\uparrow$} & \shortstack{Overlap\\@100 $\uparrow$} \\
\midrule
Global Mean & 12 & \makebox[18pt][r]{18.11}\hspace{1pt}{\fontsize{5.8}{7}\selectfont\color{Muted}(0.21)} & \makebox[18pt][r]{13.16}\hspace{1pt}{\fontsize{5.8}{7}\selectfont\color{Muted}(0.02)} & \makebox[18pt][r]{69.33}\hspace{1pt}{\fontsize{5.8}{7}\selectfont\color{Muted}(0.15)} & \makebox[18pt][r]{67.44}\hspace{1pt}{\fontsize{5.8}{7}\selectfont\color{Muted}(0.06)} & \makebox[18pt][r]{70.43}\hspace{1pt}{\fontsize{5.8}{7}\selectfont\color{Muted}(0.23)} & \makebox[18pt][r]{83.49}\hspace{1pt}{\fontsize{5.8}{7}\selectfont\color{Muted}(1.77)} & \makebox[18pt][r]{0.79}\hspace{1pt}{\fontsize{5.8}{7}\selectfont\color{Muted}(0.02)} \\
Linear$^\dagger$ & 12 & \makebox[18pt][r]{25.67}\hspace{1pt}{\fontsize{5.8}{7}\selectfont\color{Muted}(0.43)} & \makebox[18pt][r]{9.51}\hspace{1pt}{\fontsize{5.8}{7}\selectfont\color{Muted}(0.08)} & \makebox[18pt][r]{69.05}\hspace{1pt}{\fontsize{5.8}{7}\selectfont\color{Muted}(0.43)} & \makebox[18pt][r]{67.01}\hspace{1pt}{\fontsize{5.8}{7}\selectfont\color{Muted}(0.25)} & \makebox[18pt][r]{72.72}\hspace{1pt}{\fontsize{5.8}{7}\selectfont\color{Muted}(0.32)} & \makebox[18pt][r]{74.36}\hspace{1pt}{\fontsize{5.8}{7}\selectfont\color{Muted}(0.39)} & \makebox[18pt][r]{1.33}\hspace{1pt}{\fontsize{5.8}{7}\selectfont\color{Muted}(0.05)} \\
Ridge & 12 & \makebox[18pt][r]{32.67}\hspace{1pt}{\fontsize{5.8}{7}\selectfont\color{Muted}(0.43)} & \makebox[18pt][r]{6.48}\hspace{1pt}{\fontsize{5.8}{7}\selectfont\color{Muted}(0.08)} & \makebox[18pt][r]{72.47}\hspace{1pt}{\fontsize{5.8}{7}\selectfont\color{Muted}(0.14)} & \makebox[18pt][r]{70.58}\hspace{1pt}{\fontsize{5.8}{7}\selectfont\color{Muted}(0.16)} & \makebox[18pt][r]{77.76}\hspace{1pt}{\fontsize{5.8}{7}\selectfont\color{Muted}(0.06)} & \makebox[18pt][r]{82.32}\hspace{1pt}{\fontsize{5.8}{7}\selectfont\color{Muted}(0.81)} & \makebox[18pt][r]{0.98}\hspace{1pt}{\fontsize{5.8}{7}\selectfont\color{Muted}(0.06)} \\
chemCPA$^\dagger$ & 12 & \makebox[18pt][r]{26.85}\hspace{1pt}{\fontsize{5.8}{7}\selectfont\color{Muted}(0.26)} & \makebox[18pt][r]{6.99}\hspace{1pt}{\fontsize{5.8}{7}\selectfont\color{Muted}(0.05)} & \makebox[18pt][r]{71.12}\hspace{1pt}{\fontsize{5.8}{7}\selectfont\color{Muted}(0.41)} & \makebox[18pt][r]{69.76}\hspace{1pt}{\fontsize{5.8}{7}\selectfont\color{Muted}(0.46)} & \makebox[18pt][r]{76.56}\hspace{1pt}{\fontsize{5.8}{7}\selectfont\color{Muted}(0.45)} & \makebox[18pt][r]{53.39}\hspace{1pt}{\fontsize{5.8}{7}\selectfont\color{Muted}(0.84)} & \makebox[18pt][r]{15.08}\hspace{1pt}{\fontsize{5.8}{7}\selectfont\color{Muted}(0.17)} \\
PRnet & 12 & \makebox[18pt][r]{26.38}\hspace{1pt}{\fontsize{5.8}{7}\selectfont\color{Muted}(0.73)} & \makebox[18pt][r]{7.21}\hspace{1pt}{\fontsize{5.8}{7}\selectfont\color{Muted}(0.09)} & \makebox[18pt][r]{71.37}\hspace{1pt}{\fontsize{5.8}{7}\selectfont\color{Muted}(0.69)} & \makebox[18pt][r]{69.31}\hspace{1pt}{\fontsize{5.8}{7}\selectfont\color{Muted}(0.80)} & \makebox[18pt][r]{75.05}\hspace{1pt}{\fontsize{5.8}{7}\selectfont\color{Muted}(0.57)} & \makebox[18pt][r]{53.74}\hspace{1pt}{\fontsize{5.8}{7}\selectfont\color{Muted}(0.94)} & \makebox[18pt][r]{19.29}\hspace{1pt}{\fontsize{5.8}{7}\selectfont\color{Muted}(0.19)} \\
biolord & 12 & \makebox[18pt][r]{31.23}\hspace{1pt}{\fontsize{5.8}{7}\selectfont\color{Muted}(1.21)} & \makebox[18pt][r]{6.80}\hspace{1pt}{\fontsize{5.8}{7}\selectfont\color{Muted}(0.18)} & \makebox[18pt][r]{72.19}\hspace{1pt}{\fontsize{5.8}{7}\selectfont\color{Muted}(0.29)} & \makebox[18pt][r]{70.15}\hspace{1pt}{\fontsize{5.8}{7}\selectfont\color{Muted}(0.37)} & \makebox[18pt][r]{76.43}\hspace{1pt}{\fontsize{5.8}{7}\selectfont\color{Muted}(0.26)} & \makebox[18pt][r]{68.04}\hspace{1pt}{\fontsize{5.8}{7}\selectfont\color{Muted}(0.32)} & \makebox[18pt][r]{0.98}\hspace{1pt}{\fontsize{5.8}{7}\selectfont\color{Muted}(0.03)} \\
PerturbNet & 12 & \makebox[18pt][r]{26.64}\hspace{1pt}{\fontsize{5.8}{7}\selectfont\color{Muted}(0.59)} & \makebox[18pt][r]{9.34}\hspace{1pt}{\fontsize{5.8}{7}\selectfont\color{Muted}(0.27)} & \makebox[18pt][r]{70.65}\hspace{1pt}{\fontsize{5.8}{7}\selectfont\color{Muted}(0.08)} & \makebox[18pt][r]{68.38}\hspace{1pt}{\fontsize{5.8}{7}\selectfont\color{Muted}(0.12)} & \makebox[18pt][r]{74.07}\hspace{1pt}{\fontsize{5.8}{7}\selectfont\color{Muted}(0.16)} & \makebox[18pt][r]{74.53}\hspace{1pt}{\fontsize{5.8}{7}\selectfont\color{Muted}(1.16)} & \makebox[18pt][r]{1.04}\hspace{1pt}{\fontsize{5.8}{7}\selectfont\color{Muted}(0.03)} \\
CellFlow & 12 & \makebox[18pt][r]{32.88}\hspace{1pt}{\fontsize{5.8}{7}\selectfont\color{Muted}(1.10)} & \makebox[18pt][r]{6.40}\hspace{1pt}{\fontsize{5.8}{7}\selectfont\color{Muted}(0.01)} & \makebox[18pt][r]{72.03}\hspace{1pt}{\fontsize{5.8}{7}\selectfont\color{Muted}(0.28)} & \makebox[18pt][r]{70.16}\hspace{1pt}{\fontsize{5.8}{7}\selectfont\color{Muted}(0.37)} & \makebox[18pt][r]{77.34}\hspace{1pt}{\fontsize{5.8}{7}\selectfont\color{Muted}(0.21)} & \makebox[18pt][r]{84.98}\hspace{1pt}{\fontsize{5.8}{7}\selectfont\color{Muted}(0.15)} & \makebox[18pt][r]{0.69}\hspace{1pt}{\fontsize{5.8}{7}\selectfont\color{Muted}(0.03)} \\
CMonge$^\dagger$ & 12 & \makebox[18pt][r]{31.68}\hspace{1pt}{\fontsize{5.8}{7}\selectfont\color{Muted}(0.67)} & \makebox[18pt][r]{6.70}\hspace{1pt}{\fontsize{5.8}{7}\selectfont\color{Muted}(0.40)} & \makebox[18pt][r]{71.71}\hspace{1pt}{\fontsize{5.8}{7}\selectfont\color{Muted}(0.25)} & \makebox[18pt][r]{69.85}\hspace{1pt}{\fontsize{5.8}{7}\selectfont\color{Muted}(0.36)} & \makebox[18pt][r]{76.75}\hspace{1pt}{\fontsize{5.8}{7}\selectfont\color{Muted}(0.24)} & \makebox[18pt][r]{67.36}\hspace{1pt}{\fontsize{5.8}{7}\selectfont\color{Muted}(2.03)} & \makebox[18pt][r]{0.95}\hspace{1pt}{\fontsize{5.8}{7}\selectfont\color{Muted}(0.04)} \\
Prophet & 12 & \makebox[18pt][r]{29.19}\hspace{1pt}{\fontsize{5.8}{7}\selectfont\color{Muted}(0.73)} & \makebox[18pt][r]{7.04}\hspace{1pt}{\fontsize{5.8}{7}\selectfont\color{Muted}(0.09)} & \makebox[18pt][r]{72.21}\hspace{1pt}{\fontsize{5.8}{7}\selectfont\color{Muted}(0.15)} & \makebox[18pt][r]{70.02}\hspace{1pt}{\fontsize{5.8}{7}\selectfont\color{Muted}(0.11)} & \makebox[18pt][r]{76.51}\hspace{1pt}{\fontsize{5.8}{7}\selectfont\color{Muted}(0.20)} & \makebox[18pt][r]{\textbf{86.02}}\hspace{1pt}{\fontsize{5.8}{7}\selectfont\color{Muted}(0.56)} & \makebox[18pt][r]{0.86}\hspace{1pt}{\fontsize{5.8}{7}\selectfont\color{Muted}(0.14)} \\
PrePR-CT & 12 & \makebox[18pt][r]{28.24}\hspace{1pt}{\fontsize{5.8}{7}\selectfont\color{Muted}(0.85)} & \makebox[18pt][r]{7.59}\hspace{1pt}{\fontsize{5.8}{7}\selectfont\color{Muted}(0.74)} & \makebox[18pt][r]{72.04}\hspace{1pt}{\fontsize{5.8}{7}\selectfont\color{Muted}(0.66)} & \makebox[18pt][r]{69.76}\hspace{1pt}{\fontsize{5.8}{7}\selectfont\color{Muted}(0.61)} & \makebox[18pt][r]{75.21}\hspace{1pt}{\fontsize{5.8}{7}\selectfont\color{Muted}(0.51)} & \makebox[18pt][r]{75.59}\hspace{1pt}{\fontsize{5.8}{7}\selectfont\color{Muted}(4.29)} & \makebox[18pt][r]{0.97}\hspace{1pt}{\fontsize{5.8}{7}\selectfont\color{Muted}(0.06)} \\
XPert$^*$ & 7 & \makebox[18pt][r]{\textcolor{Muted}{39.00}}\hspace{1pt}{\fontsize{5.8}{7}\selectfont\color{Muted}(0.08)} & \makebox[18pt][r]{\textcolor{Muted}{7.40}}\hspace{1pt}{\fontsize{5.8}{7}\selectfont\color{Muted}(0.32)} & \makebox[18pt][r]{\textcolor{Muted}{64.00}}\hspace{1pt}{\fontsize{5.8}{7}\selectfont\color{Muted}(1.37)} & \makebox[18pt][r]{\textcolor{Muted}{63.74}}\hspace{1pt}{\fontsize{5.8}{7}\selectfont\color{Muted}(1.34)} & \makebox[18pt][r]{\textcolor{Muted}{74.24}}\hspace{1pt}{\fontsize{5.8}{7}\selectfont\color{Muted}(2.66)} & \makebox[18pt][r]{\textcolor{Muted}{76.12}}\hspace{1pt}{\fontsize{5.8}{7}\selectfont\color{Muted}(2.06)} & \makebox[18pt][r]{\textcolor{Muted}{0.36}}\hspace{1pt}{\fontsize{5.8}{7}\selectfont\color{Muted}(0.05)} \\
State & 12 & \makebox[18pt][r]{35.54}\hspace{1pt}{\fontsize{5.8}{7}\selectfont\color{Muted}(0.41)} & \makebox[18pt][r]{6.05}\hspace{1pt}{\fontsize{5.8}{7}\selectfont\color{Muted}(0.09)} & \makebox[18pt][r]{76.73}\hspace{1pt}{\fontsize{5.8}{7}\selectfont\color{Muted}(0.20)} & \makebox[18pt][r]{74.89}\hspace{1pt}{\fontsize{5.8}{7}\selectfont\color{Muted}(0.21)} & \makebox[18pt][r]{84.04}\hspace{1pt}{\fontsize{5.8}{7}\selectfont\color{Muted}(0.61)} & \makebox[18pt][r]{55.83}\hspace{1pt}{\fontsize{5.8}{7}\selectfont\color{Muted}(1.02)} & \makebox[18pt][r]{\textbf{19.50}}\hspace{1pt}{\fontsize{5.8}{7}\selectfont\color{Muted}(0.16)} \\
\midrule
Bison & 12 & \makebox[18pt][r]{\textbf{43.37}}\hspace{1pt}{\fontsize{5.8}{7}\selectfont\color{Muted}(0.55)} & \makebox[18pt][r]{\textbf{5.68}}\hspace{1pt}{\fontsize{5.8}{7}\selectfont\color{Muted}(0.08)} & \makebox[18pt][r]{\textbf{77.82}}\hspace{1pt}{\fontsize{5.8}{7}\selectfont\color{Muted}(0.48)} & \makebox[18pt][r]{\textbf{76.47}}\hspace{1pt}{\fontsize{5.8}{7}\selectfont\color{Muted}(0.52)} & \makebox[18pt][r]{\textbf{84.58}}\hspace{1pt}{\fontsize{5.8}{7}\selectfont\color{Muted}(0.38)} & \makebox[18pt][r]{70.06}\hspace{1pt}{\fontsize{5.8}{7}\selectfont\color{Muted}(0.72)} & \makebox[18pt][r]{16.70}\hspace{1pt}{\fontsize{5.8}{7}\selectfont\color{Muted}(1.09)} \\
\bottomrule
\end{tabular*}
\endgroup
\end{table}

\subsection{Joint Training Improves Unseen-Compound Prediction}\label{sec:joint_gains}

We vary each DLM between dataset-specific and joint training, fixing architectures, objectives, tokenizer and inference (Table~\ref{tab:joint_solo}; Appendix~\ref{app:joint_solo_details}).

\noindent
\begin{minipage}[t]{0.47\textwidth}
\vspace{0pt}
\begingroup
\makeatletter\def\@captype{figure}\makeatother
\centering
\setlength{\abovecaptionskip}{3pt}
\includegraphics[width=\linewidth]{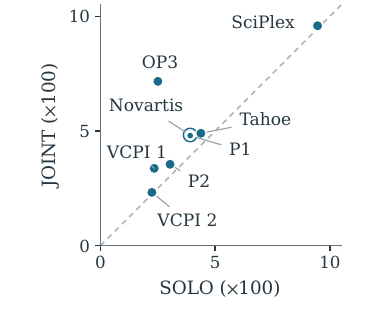}
\caption{\raggedright\textbf{JOINT--SOLO drug contrast.} Native-view TEST means; dashed line: equal scores. P1/P2: L1000.}
\label{fig:joint_solo_scatter}
\endgroup

\end{minipage}\hfill
\begin{minipage}[t]{0.50\textwidth}
\vspace{0pt}
\begingroup
\makeatletter
\def\@captype{table}
\makeatother
\small
\setlength{\abovecaptionskip}{0pt}
\setlength{\belowcaptionskip}{3pt}
\caption{\textbf{Joint training by component.}
TEST Pearson ($\times100$), averaged over views.
S/J: separate/joint training.
Overall: context/molecular; drug: molecular (joint context).
Macro: equal dataset weights.}
\label{tab:joint_solo}
\centering
\setlength{\tabcolsep}{1.5pt}
\begin{tabular*}{\linewidth}{@{\extracolsep{\fill}}lrrrrrr@{}}
\toprule
& \multicolumn{4}{c}{Overall response} & \multicolumn{2}{c}{Drug contrast} \\
\cmidrule(lr){2-5}\cmidrule(lr){6-7}
Dataset & S/S & S/J & J/S & \textbf{J/J} & S & \textbf{J} \\
\midrule
OP3 & 39.66 & 39.41 & 40.44 & \textbf{40.10} & 2.50 & \textbf{7.16} \\
SciPlex & 45.36 & 46.25 & 46.37 & \textbf{47.18} & 9.46 & \textbf{9.58} \\
VCPI 1 & 33.36 & 33.33 & 33.27 & \textbf{33.50} & 2.34 & \textbf{3.37} \\
VCPI 2 & 33.46 & 33.31 & 33.52 & \textbf{33.61} & 2.24 & \textbf{2.32} \\
Novartis & 49.71 & 50.48 & 52.11 & \textbf{52.60} & 3.91 & \textbf{4.82} \\
Tahoe & 53.73 & 53.84 & 54.91 & \textbf{54.86} & 4.38 & \textbf{4.90} \\
L1000 P1 & 36.37 & 36.50 & 35.97 & \textbf{36.20} & 3.91 & \textbf{4.80} \\
L1000 P2 & 21.20 & 21.44 & 20.77 & \textbf{21.21} & 3.04 & \textbf{3.55} \\
\midrule
Macro & 39.11 & 39.32 & 39.67 & \textbf{39.91} & 3.97 & \textbf{5.06} \\
\bottomrule
\end{tabular*}
\endgroup

\end{minipage}
\par\smallskip

Joint training improves mean overall response by 2.0\% and mean drug contrast by 27.4\%.
The mixed configurations show complementary effects: joint context-response training provides most of the overall gain, while joint molecular-effect training raises drug contrast from 0.0397 to 0.0506.
Combining both gives the highest overall macro, 0.3991.

Drug contrast improves in all eight dataset means (Figure~\ref{fig:joint_solo_scatter}), while overall response improves in seven.
On OP3, drug contrast nearly triples (0.0250 to 0.0716), while overall response changes only slightly.
SciPlex lies near the diagonal at a higher absolute score, illustrating the distinction between strong prediction and a large joint-training gain.
These patterns show why both metrics are needed: better prediction of molecular differences can yield only a modest change in overall agreement.
On L1000 P1, the drug-contrast gain coexists with a small decrease in overall response.

\subsection{Interpreting Cross-Dataset Response Prediction}\label{sec:interpretability}

\begin{figure}[!htb]
\centering
\includegraphics[width=\textwidth]{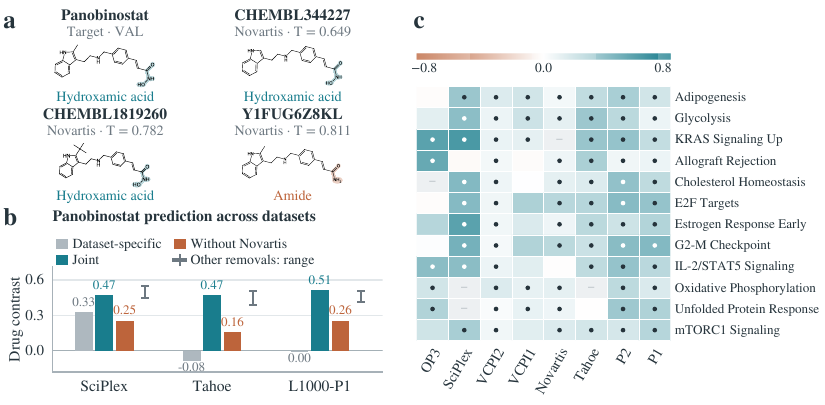}
\caption{\textbf{Molecular evidence and pathway signals.} (a) Panobinostat and Novartis TRAIN analogs ($T$: Tanimoto similarity). (b) Panobinostat VAL drug contrast; whiskers span other source removals. (c) Twelve Hallmark pathways predicted in at least six datasets. Color: truth-SD-scaled HVG TEST Pearson; dots: pooled VAL/TEST FDR $<0.10$ and positive correlation in both splits. Gray cells indicate unavailable values; P1/P2 denote L1000 phases.}
\label{fig:interpretability_combined}
\label{fig:panobinostat_case}
\label{fig:pathway_map}
\end{figure}

Panobinostat illustrates how another screen can supply useful molecular supervision (Figure~\ref{fig:interpretability_combined}a,b).
Its closest TRAIN analogs (Tanimoto 0.65--0.81) occur in Novartis and are absent from the three target datasets' training data; two retain its hydroxamic acid.
Joint training improves Panobinostat prediction in SciPlex, Tahoe and L1000 P1, whereas removing Novartis gives the lowest score among all source removals in each.
The case links structural similarity in the source screen to a measurable prediction benefit in three distinct target settings.

\begin{samepage}
Across unseen compounds, the predicted molecular differences also organize into biological pathways.
An exploratory Hallmark analysis identifies 41 pathways with significant molecule-level predictions in at least two datasets.
The twelve found in at least six datasets span proliferation, stress and inflammatory signaling, and metabolism (Figure~\ref{fig:interpretability_combined}c).
These associations show that predicted differences between compounds extend to coordinated cellular programs across multiple native gene panels.
Full source-removal, neighbor-copying and pathway results are in Appendix~\ref{app:interpretability}.
\par\end{samepage}

\subsection{Ablations}\label{sec:ablation_overview}
The ablations support separating the shared response from molecular differences and modeling the cellular context of those differences.
Adding the molecular correction improves overall prediction, while moderate scaling outperforms the unscaled deviation.
Context modulation and the interaction loss provide most of the drug-contrast gain, with a smaller increment from within-plate contrast supervision.
The full component and scaling comparisons are in Appendix~\ref{app:ablations}.

\section{Conclusion}
\label{sec:conclusion}

Native gene panels can be preserved while chemical-response learning is shared across datasets. Bison demonstrates that a common gene interface lets complementary screens contribute to unseen-compound prediction, turning fragmented measurements into usable molecular supervision. Joint training improves both overall response agreement and prediction of drug-specific differences, and matched contrasts make those differences explicit in both the learning objective and the evaluation. These findings suggest that integrating perturbation datasets is as much a question of aligning the prediction task as of combining measurements. Retaining native measurements while sharing molecular learning can bring diverse chemical screens into a unified model.

\clearpage
\section*{Reproducibility Statement}
The code for reproducing our experiments will be provided in the supplementary files.
Section~\ref{sec:benchmark} and Appendix~\ref{app:benchmark_data} describe data sources, native gene panels, compound splits, control matching, and metrics.
Section~\ref{sec:methods} details the model objectives and inference; Appendix~\ref{app:baseline_adaptations} documents training, checkpoint selection and baseline adaptations; Appendix~\ref{app:experiment_details} provides the complete results.
The main benchmark reports means and sample standard deviations over three independent global compound splits.

\section*{AI Use Statement}
Generative AI tools were used for manuscript drafting, revision, and LaTeX formatting, as well as experiment implementation and analysis.
The authors take responsibility for the final manuscript, experimental results, and conclusions.

\let\section\bisonoriginalsection
\let\subsection\bisonoriginalsubsection
\bibliography{references}
\bibliographystyle{iclr2027_conference}
\clearpage
\appendix
\raggedbottom
\section{Data and Evaluation}
\label{app:benchmark_data}

\subsection{Dataset Coverage and Gene Panels}
\label{app:data_coverage}

Table~\ref{tab:benchmark_data} lists the native coverage of the eight datasets.
The six count-based datasets each provide Full and HVG views; each L1000 dataset contributes one landmark view to the fourteen-view benchmark.
Compound assignments, native panels and held-out records are shared by all methods.
As described in Section~\ref{sec:benchmark_protocol}, the released HVG panels are constructed from all chemically eligible profiles, whereas model fitting and reference estimation use TRAIN data.

\begin{table}[H]
\centering
\caption{Benchmark coverage. Profiles include treated and control bulk/pseudobulk records. Full counts measured genes in the benchmark vocabulary; the two L1000 datasets each contribute one 978-gene landmark view. Molecules and cellular contexts are deduplicated in the total row.}
\label{tab:benchmark_data}
\small
\setlength{\tabcolsep}{7pt}
\begin{tabular}{lrrrrr}
\toprule
Dataset & Profiles & Molecules & Contexts & Full genes & HVG/landmark \\
\midrule
OP3 & 1,813 & 138 & 4 & 5,198 & 2,000 \\
SciPlex & 4,974 & 186 & 3 & 20,251 & 2,000 \\
VCPI1 & 27,517 & 2,272 & 1 & 20,629 & 2,000 \\
VCPI2 & 18,139 & 1,488 & 1 & 20,629 & 2,000 \\
Novartis & 46,748 & 3,770 & 1 & 20,038 & 2,000 \\
Tahoe & 67,018 & 376 & 50 & 20,573 & 2,000 \\
L1000 P1 & 692,787 & 9,233 & 70 & --- & 978 \\
L1000 P2 & 333,263 & 1,760 & 30 & --- & 978 \\
\midrule
Total / unique & 1,192,259 & 16,771 & 131 & \multicolumn{2}{c}{14 views} \\
\bottomrule
\end{tabular}
\end{table}

\paragraph{Chemical-space visualization.}
Figure~\ref{fig:chemical_space} uses all 16,771 treated compounds across TRAIN, VAL and TEST.
We compute 2,048-bit Morgan fingerprints (radius 2, without chirality) from standardized connectivity SMILES, reduce them to 64 dimensions by truncated SVD, and fit one two-dimensional t-SNE embedding with perplexity 50 and 750 iterations.
Each panel highlights one dataset on the same embedding; compounds shared across sources appear in every applicable panel.
These coordinates are used only for visualization.

\subsection{Scoring and Aggregation}
\label{app:scoring}

\paragraph{Control matching.}\label{app:control_reference}
The overall-response scorer subtracts TRAIN controls matched by context, plate and time from each treated record and its prediction, then averages the deltas by context--compound--dose--time condition before computing correlation across genes.
This preserves record-level control pairing when a condition spans several plates.
In L1000 Phase~1, records without exact matches fall back to context--time and then context TRAIN controls; all TEST records in the first split have exact matches.
The observed scoring reference is distinct from the empirical-Bayes reference used as model input.
Pooling controls across plates adds a plate-to-context offset to the within-plate response:
\begin{equation}
\mathbf x-\mathbf C_{c,t}
=\underbrace{\mathbf x-\mathbf C_{c,p,t}}_{\text{within-plate response}}
+\underbrace{\mathbf C_{c,p,t}-\mathbf C_{c,t}}_{\text{plate-to-context offset}},
\label{eq:control_reference_decomposition}
\end{equation}
where $\mathbf C_{c,p,t}$ and $\mathbf C_{c,t}$ are plate-matched and pooled context--time TRAIN control means.
Appendix~\ref{app:reference_sensitivity} compares these references using identical predictions.

\paragraph{Drug contrast.}
Equation~\ref{eq:drug_contrast_score} uses groups with at least three distinct compounds at the same context, plate, dose and time.
Replicates are averaged per compound before observed and predicted responses are centered separately across the group's compounds.
The score averages finite across-gene correlations over group--compound pairs.
Centering cancels the additive response and estimated control error shared within a group.
Constant predictions have zero contrast and an undefined correlation, reported as NA.
Only records with exact TRAIN control matches enter this score; L1000 Phase~1 excludes 79 TEST rows in the second split and 67 in the third.
The contrast evaluator restores its FP16-exported observations and controls to FP32 and clips predictions at zero, including L1000.
Overall-response evaluation follows the native scorer's expression policy.

\paragraph{Cell-Eval metrics.}
We use Cell-Eval 0.8.2 (commit \texttt{f9e4c058}), its full metric profile, and the plate-matched recombined-replicate protocol.
Each replicate is normalized with its exact TRAIN plate-control mean before biological replicates are recombined.
MAE $\Delta$ and MSE $\Delta$ measure absolute and squared response errors.
Perturbation discrimination (PDS) ranks a predicted condition effect against observed candidates in the same context--time stratum, returning one minus the zero-based rank divided by the number of candidates.
Candidates can span plates and doses; L1/L2 distances depend on magnitude, whereas cosine emphasizes direction.
DE direction measures sign agreement on observed significant genes, and log-fold-change Spearman measures their effect ordering.
Overlap@100 compares significant-gene lists ranked by absolute log-fold change, using at most 100 genes and the observed list length as denominator.
The significance threshold is FDR 0.05.
DE-count Spearman correlates the numbers of significant genes across conditions; the full profile also reports DE recall, ROC/PR-AUC, precision, E-distance and clustering agreement.
Cell-Eval supports the twelve count-based views; negative L1000 Level-3 profiles fall outside its input domain.
Condition-coverage exceptions are listed in Appendix~\ref{app:baseline_missing}.

\paragraph{Macro averages.}
The main benchmark weights fourteen native views equally.
Training-regime, coefficient and source-removal comparisons first average views within each dataset and then weight the eight datasets equally.
Cell-Eval averages a fixed common set of defined views for each metric: twelve for Pearson, error and discrimination scores, and ten for DE direction.
Three-split summaries report the mean and sample standard deviation of split-level macros.
Incomplete-coverage macros use a fixed available-view set across splits, are marked with $*$, and are excluded from full-coverage ranking.

\subsection{Repeatability Reference}
\label{app:repeatability}

We compare each measured drug contrast with repeats of the same context--compound--dose--time condition on other eligible plates, averaging the pairwise correlations for that target.
Bison is evaluated against the same finite repeat-eligible targets.
VCPI1/2 have no eligible cross-plate repeats: we instead use the first two distinct wells per compound, retain groups with at least three paired compounds, and center the two well sets separately.
Their reference and model scores therefore concern single-well targets, averaged over the two wells.
Table~\ref{tab:repeatability_reference} reports the resulting cohort sizes and correlations; Tahoe's repeat cohort covers about one quarter of its full drug-contrast cohort.

Within each view and split, the repeat-relative percentage is $100\,\bar r_{\mathrm{model}}/\bar r_{\mathrm{repeat}}$ on the matched cohort.
We average these percentages over views within datasets, equally over datasets, and then across splits, giving $45.0\%$ for Bison.
This is agreement relative to another noisy measurement, rather than a hard prediction ceiling.

\begin{table}[!htbp]
\centering
\caption{\textbf{Drug-contrast agreement relative to repeated measurements.} Correlations and ratios are means across three splits. Unit counts span native views and splits; VCPI1/2 use paired wells, and other datasets use cross-plate repeats.}
\label{tab:repeatability_reference}
\label{tab:contrast_noise_ceiling}
\small
\setlength{\tabcolsep}{8pt}
\begin{tabular}{lrrrr}
\toprule
Dataset & Eligible units & Repeat $r$ & Bison $r$ & Bison / repeat (\%) \\
\midrule
OP3 & 268--353 & 0.228 & 0.091 & 40.2 \\
SciPlex & 870--944 & 0.147 & 0.078 & 53.1 \\
VCPI1 & 2,498--2,692 & 0.030 & 0.024 & 80.2 \\
VCPI2 & 1,610--1,723 & 0.030 & 0.016 & 53.0 \\
Novartis & 9,276--9,430 & 0.284 & 0.039 & 13.9 \\
Tahoe & 3,010--3,492 & 0.298 & 0.056 & 18.5 \\
L1000 P1 & 93,191--101,773 & 0.086 & 0.050 & 57.8 \\
L1000 P2 & 58,217--67,904 & 0.098 & 0.042 & 42.9 \\
\midrule
Equal-dataset mean & --- & 0.150 & 0.049 & 45.0 \\
\bottomrule
\end{tabular}
\par\vspace{3pt}
\begin{minipage}{0.94\textwidth}\footnotesize
Ratios are computed within each view and split before averaging views, datasets and splits; the final column is not the ratio of the two displayed aggregate correlations. The eight-dataset mean is $45.0\%$ with sample SD $2.5$ percentage points across splits.
\end{minipage}
\end{table}

\section{Implementation and Training}
\label{app:baseline_adaptations}

\subsection{Bison Architecture and Training}
\label{app:bison_training}
\label{app:training_evaluation_details}

\paragraph{Shared representation.}
The tokenizer uses 128 slots and FSQ levels $(8,8,8,6,5)$.
Learned slot queries aggregate gene identity--value pairs, and native-gene queries decode the requested panel.
Joint pretraining minimizes scaled residual reconstruction with a coordinate-saturation penalty; the tokenizer is then frozen for both DLMs.
Each compound split uses its own tokenizer and TRAIN-derived scales and reference statistics.

\paragraph{DLM architecture and conditioning.}
Both DLMs have width 256, four Transformer blocks, eight attention heads and a rank-32 chemical field in their digit logits.
The molecular-effect DLM additionally uses four chemical memory tokens formed from reference-centered Morgan fingerprints and dose/time features.
A 64-dimensional embedding identifies each dataset-specific cellular context.
Zero-initialized projections supply token-wise scales and shifts and an additive scalar-memory contribution; shifts are gated off for control or dropped chemical conditions.
TRAIN controls provide the empirical-Bayes reference, plate--context offsets, control counts, variance and matching status, while a view embedding identifies the native panel.
Plate information enters through these measured features, without a learned plate-identity embedding.

\paragraph{Optimization and checkpoint selection.}
Both branches use square-root dataset quotas and AdamW.
The context-response branch minimizes token and decoded-response losses with weights 1 and 2.5, using a 2,936-update warmup and cosine decay defined through 88,155 updates.
Its checkpoint criterion is soft-decoded VAL overall response, averaged over native views within each of seven datasets and then across datasets, excluding L1000 Phase~1.
The selected context-response checkpoints are 80,000, 75,000 and 35,000 updates on the three compound splits.
This branch is trained and selected for complete-response prediction; composition uses its vehicle-conditioned output.

The molecular-effect branch uses peak learning rate $3\times10^{-4}$, floor $3\times10^{-5}$, 100 warmup updates, weight decay 0.05 and gradient clipping at 1.
Its cosine schedule is defined over 52,893 updates, with training ending after the last complete VAL evaluation at 50,000.
Token, response, within-plate contrast and context-interaction losses have weights 1, 2.5, 10 and 10.
Complete VAL evaluations every 5,000 updates select drug-contrast correlation, averaging native views within datasets and then all eight datasets equally.
The selected molecular checkpoints are 40,000, 25,000 and 30,000 updates for the main benchmark.

\paragraph{Loss sampling.}
Token cross-entropy is averaged over masked slots and FSQ digits.
Decoded losses use fully masked inputs, restore expression units, clip at zero, subtract matched TRAIN controls and aggregate replicates before computing $1-\rho$.
Gradients pass through the frozen decoder.
Within-plate contrast samples up to six compounds from groups containing at least three, centering truth and prediction over the same compounds.
Context-interaction supervision samples complete grids of up to four compounds and four contexts under matched plate, dose and time, requiring at least two of each.
Only OP3 and Tahoe supply eligible grids; both native views contribute to the losses.

\subsection{Prediction and Calibration}
\label{app:bison_prediction}

\paragraph{Reference calls.}
The fixed molecular reference $m_0$ is the observed TRAIN-control compound of each dataset, represented by its vehicle fingerprint.
For the shared-response call $\mathbf r_\eta(m_0,c,G)$, dose and treatment-status scalars retain the query values, while the reference-centered chemical field is zero.
For the molecular reference $\mathbf r_\theta(m_0,c_0,G)$, dose and status use control values, with matched controls, time, gene view and cellular-context embedding retained.
The molecular query uses the requested drug and dose.
These calls are composed with the fixed $\alpha=0.25$ in Equation~\ref{eq:composed_prediction}.

\paragraph{One-step soft decoding.}
For either DLM $\omega\in\{\eta,\theta\}$, slot $\ell$ and FSQ coordinate $k$, inference computes
\begin{equation}
\bar z^\omega_{\ell k}=\sum_{j=0}^{L_k-1}p_\omega(q_{\ell k}=j\mid\mathrm{MASK},m,c,G)\,\zeta_k(j),\qquad
\mathbf r_\omega(m,c,G)=a_d^{-1}D_\psi(\bar{\mathbf z}^\omega,G),
\label{eq:soft_decode}
\end{equation}
where $\zeta_k(j)=(j-\lfloor L_k/2\rfloor)/\lfloor L_k/2\rfloor$ is the normalized digit coordinate.
Each call starts with all slots masked and uses one conditional forward pass and one decode.
Reference encoding, DLM computation and digit heads use FP32; the frozen decoder uses BF16 and decodes one condition row at a time.
The three decoded residuals are combined before reference restoration and calibration.

\paragraph{Amplitude estimation.}\label{app:calibration_details}
For each dataset, we sample at most 6,000 TRAIN treated rows without replacement, subtract exact TRAIN plate-control means, and average replicates within context--condition--time groups.
Their observed and predicted effect matrices yield one amplitude scalar per dataset and view through Equation~\ref{eq:response_calibration}, with $\epsilon=10^{-9}$ and $\|\cdot\|_1$ denoting the sum of absolute entries.
Calibration scales the complete untruncated response around the exact control mean before nonnegative clipping.

\paragraph{Detection-aware zeros.}
For TRAIN detection frequency $p_{d,g}$, the final output is
\begin{equation}
 x^{\mathrm{post}}_{i,g}=\begin{cases}
 0,&x^{\mathrm{cal}}_{i,g}<\tau_d\ \text{and}\ p_{d,g}<0.1,\\
 x^{\mathrm{cal}}_{i,g},&\text{otherwise}.
 \end{cases}
\label{eq:detection_zeros}
\end{equation}
Detection frequencies use the TRAIN treated subset with exact control matches.
The threshold $\tau_d$ is half the median per-row minimum positive expression in a separate sample of at most 3,000 TRAIN treated rows, ignoring all-zero rows.
It uses the complete native panel and is shared by Full and HVG; detection frequencies are restricted to the queried genes.
The rule and 0.1 cutoff were selected on the first split's VAL set and fixed across splits, with numerical scales, thresholds and frequencies estimated from each split's TRAIN data.
The count-based datasets use both steps; L1000 retains the uncalibrated composition with nonnegative clipping.
Appendix~\ref{app:calibration_results} reports their effects on the evaluation metrics.

\subsection{Benchmark Execution}
\label{app:baseline_execution}

\paragraph{Data and fitting.}
We use released neural-model implementations and fit each baseline separately for each dataset and native gene view, using the prepared bulk/pseudobulk profiles and global compound assignments.
TRAIN contains the training compounds and available controls.
Feature scalers, expression projections and control-derived graphs are fitted on TRAIN; molecular encoders also process held-out structures to construct their input features.
Distributional models train on collections of aggregate profiles from the relevant experimental groups.

\paragraph{Training and prediction.}
Each baseline retains its objective and optimization schedule, selecting checkpoints on VAL compounds or using the terminal checkpoint for fixed-schedule runs.
Adapters export post-perturbation expression with the original observation and ordered gene identifiers.
Latent predictions are decoded to the native panel; generated or transported profiles are averaged when the adapter returns a condition mean, then assigned to the corresponding held-out records.
The common evaluator applies control subtraction, replicate aggregation and scoring.
Supplementary run configurations record optimization settings, source revisions, selected checkpoints and prediction identifiers.

\subsection{Baseline Adaptations}
\label{app:baseline_models}

\paragraph{Global Mean and Linear.}
Global Mean predicts the row-weighted mean of TRAIN treated profiles in the dataset and gene view.
Linear fits absolute expression by unregularized minimum-norm least squares with an intercept, a 1,024-bit radius-two feature Morgan fingerprint, log-transformed dose and time, and TRAIN-context indicators.
Duplicate inputs retain their multiplicity as fitting weights.
Its FP64 singular-value cutoff is determined by machine precision and matrix size.

\paragraph{Ridge.} The \texttt{ridge\_morgan} configuration uses multi-output ridge regression on condition-mean expression deltas.
Its chemical input is a 1,024-bit radius-2 feature Morgan fingerprint (FCFP4), multiplied by $\log(1+\mathrm{dose}_{\mu\mathrm M})$, alongside $\log(1+\mathrm{dose}_{\mu\mathrm M})$, an intercept, and context/time indicators.
Targets subtract context/time-matched TRAIN control means, with a same-context fallback.
The regularization coefficient is selected by validation delta correlation.
Adding the matched control mean reconstructs the predicted expression profile.

\paragraph{chemCPA~\citep{hetzel2022chemcpa}.} We train the released compositional autoencoder from scratch on each native panel, using RDKit2D descriptors, its learned dose scaler, and cellular context/time covariates.
The encoder, decoder and compositional training objective are retained.
At inference, the model applies the requested perturbation to context/time-matched TRAIN control profiles, with a same-context fallback, and averages the decoded responses.
The benchmark uses dataset-specific training in place of the original L1000-pretraining pipeline.

\paragraph{PRnet~\citep{qi2024prnet}.} We retain the released prediction network, training loss and 1,024-bit FCFP4 chemical representation.
Each treated profile is paired with a TRAIN control from the same context and time, falling back to the same context when needed.
Dose follows the dataset-specific transformation in the released interface.
The model predicts the post-perturbation profile directly on the benchmark gene panel.

\paragraph{biolord~\citep{piran2024biolord}.} We retain the released expression decoder and disentangled latent/attribute model.
Chemical attributes use RDKit2D descriptors, with low-variance columns removed and scaling fitted on TRAIN compounds; dose and time are appended as continuous attributes.
Prediction combines learned unknown latents from context/time-matched TRAIN controls with the requested chemical, dose and time attributes, then averages the decoded profiles for each target condition.
Missing exact-time controls use the same-context pool.

\paragraph{PerturbNet~\citep{yu2025perturbnet}.} The \texttt{perturbnet\_morgan} configuration uses the released Gaussian expression-VAE branch and conditional invertible flow.
We replace the pretrained chemical-VAE representation with 1,024-bit radius-2 Morgan fingerprints to cover the benchmark compounds, and condition the flow on chemical features, context, dose and time.
The expression VAE fits TRAIN profiles, followed by flow training on treated TRAIN profiles with the native likelihood objective.
Inference samples expression latents and averages their decoded means.
The number of samples equals the size of the matched TRAIN control pool; control expression values are not inputs to this conditional generator.

\paragraph{CellFlow~\citep{klein2025cellflow}.} We retain the official conditional flow-matching model and fit its expression PCA on TRAIN profiles.
Conditioning uses Morgan fingerprints, dose, time and cellular context.
Source and target distributions consist of the available aggregate control and treated profiles.
For a requested condition, the model transports context/time-matched TRAIN controls, with same-context fallback, and averages the transported profiles after reconstruction in the native gene space.
Validation delta correlation selects the checkpoint.

\paragraph{CMonge~\citep{driessen2026cmonge}.} We retain the released autoencoder, conditional transport network, Sinkhorn fitting loss and Monge-gap regularizer.
Target distributions contain aggregate profiles from each context--condition group; source distributions use context/time-matched TRAIN controls with same-context fallback.
Small groups are sampled with replacement to form training minibatches.
RDKit2D descriptors and dose condition the map; time enters through group construction and control matching.
Decoded transported profiles are averaged for each condition.
The reported L1000 Phase~1 compatibility configuration uses $\log(1+\mathrm{dose}_{\mathrm{nM}})$ in place of $\log(\mathrm{dose}_{\mathrm{nM}})$ to include chemically labelled zero-dose records.

\paragraph{Prophet~\citep{ji2024prophet}.} The \texttt{prophet\_dt} configuration retains the released scalar gene-readout Transformer and MSE objective.
Context representations are constructed from TRAIN control expression using the authors' new-context procedure.
We resize the learned gene-readout embedding table to each native panel and append standardized $\log(1+\mathrm{dose}_{\mu\mathrm M})$ and $\log(1+\mathrm{time}_{\mathrm h})$ to the released 1,219-dimensional chemical representation.
The intervention input layer is widened accordingly.
Predicted scalar readouts are inverse-transformed and assembled into the original row and gene order.

\paragraph{PrePR-CT~\citep{alsulami2026preprct}.} The \texttt{preprct\_dt} configuration follows the published bulk-data branch.
Context-specific gene graphs use correlations among TRAIN controls, with control means and variances as node features.
We supply aggregate expression profiles directly and retain the graph encoder, predictor and grouped Earth Mover's Distance objective.
The released 124 RDKit descriptors are extended with standardized log-dose and log-time features, widening the first perturbation layer from $124\!\to\!124$ to $126\!\to\!124$.
TRAIN controls paired by context/time provide the basal expression inputs, with same-context fallback.

\paragraph{XPert~\citep{guo2026xpert}.} We retain the official attention architecture, four-component loss, Uni-Mol features, heterogeneous-graph drug features, and dose/time tokens.
Basal profiles are estimated from TRAIN controls using the most specific available context/plate/time match.
Released gene embeddings are reused where available; out-of-vocabulary genes receive seeded trainable vectors.
Compounds absent from the heterogeneous graph use its mean drug embedding together with their own Uni-Mol representation.
Predictions follow the native gene list, and validation delta correlation selects the checkpoint.

\paragraph{State~\citep{adduri2025state}.} We retain the released transition backbone, decoder and training objective.
Its perturbation input is replaced with a 1,024-bit Morgan fingerprint plus normalized log-dose and log-time features.
Each aggregate expression profile forms a singleton set (set size one).
TRAIN controls are preferentially matched by context and plate, with same-context fallback.

\subsection{Run and Metric Availability}
\label{app:baseline_missing}

\paragraph{Resource limits.}
OOM and OOT denote memory and runtime limits, respectively.
XPert's dense attention grows quadratically with gene count, and its larger Full-panel runs exceed the memory budget; OP3 Full and HVG/landmark runs completed.
CellFlow integrates its learned flow over matched controls for every condition, making the broad compound--context coverage of L1000 costly; these two runs exceed the runtime budget.
Available views remain in the per-view tables, while partial macros are marked and excluded from complete-coverage ranking.
The third-split L1000 Phase~1 HVG chemCPA run completed 240 of its maximum 300 epochs and uses the best checkpoint among those completed VAL evaluations.

\paragraph{Cell-Eval coverage.}
Some baseline outputs fail Cell-Eval's input-scale validation, and their official reruns exclude invalid condition groups.
This affects thirteen method--view--split cells: CMonge on both views of OP3, Novartis and Tahoe in the second split and Novartis in the third; chemCPA on Tahoe Full in the second; and Linear on both VCPI2 views in the first and third.
Table~\ref{tab:official_metrics} marks these three methods because their condition coverage differs; Bison uses all conditions.
XPert covers six HVG views and OP3 Full under Cell-Eval and is excluded from complete-coverage ranking.
The native benchmark uses the complete original prediction sets.
Fixed view sets and undefined drug-contrast entries are described in Appendix~\ref{app:multiseed_benchmark}.

\section{Additional Results}
\label{app:experiment_details}

\subsection{Complete Benchmark Results}
\label{app:multiseed_benchmark}

Tables~\ref{tab:benchmark_results_views} and~\ref{tab:benchmark_drug_contrast} report overall-response and drug-contrast Pearson for every native view across three independent global compound partitions.
Each cell gives the mean and sample standard deviation across split-and-training runs.
Bison holds its training initialization fixed and retrains with each split-specific tokenizer; stochastic benchmark methods also vary initialization across partitions.
Global Mean and Linear are deterministic fits.
The overall-response table uses the original benchmark's native \texttt{pearson\_delta\_batch} scorer, including its TRAIN-only control fallback.
Drug contrast is computed from the same runs and held-out predictions, with the exact-control protocol in Appendix~\ref{app:scoring}.

Macros average the same available views within each split before taking the three-split mean and SD.
Nine benchmark families and both simple baselines provide predictions for all fourteen views; Global Mean's drug contrast is undefined because it predicts no differences between compounds.
CellFlow covers twelve views.
XPert covers nine for overall response and eight for drug contrast, since VCPI1 HVG contrast is undefined in the second split.
Their partial macros are marked and excluded from complete-coverage ranking.

\begin{table}[H]
\centering
\caption{\textbf{Overall response on globally unseen compounds.} Native plate-matched Pearson $\Delta\times100$, mean (sample SD) across three compound splits. Full/HVG rows are blue/orange and ranked separately; P1/P2 denote L1000 Phases 1/2, each with one landmark view. Bold and underline mark the best and second-best displayed means, respectively.}
\label{tab:benchmark_results_views}
\begingroup
\definecolor{BenchFull}{HTML}{2E6185}
\definecolor{BenchHVG}{HTML}{9C5D32}
\definecolor{BenchFullBg}{HTML}{EDF3F8}
\definecolor{BenchHVGBg}{HTML}{F8F0E9}
\fontsize{8}{9.6}\selectfont
\setlength{\tabcolsep}{1.0pt}
\renewcommand{\arraystretch}{1.02}
\newcommand{\benchstat}[3]{\makebox[16pt][r]{#1}\hspace{1pt}\makebox[15pt][r]{{\fontsize{5.8}{7}\selectfont\color{black!62}(#2)}}\makebox[2pt][l]{\textsuperscript{#3}}}
\newcommand{\benchsingle}[2]{\makebox[16pt][r]{#1}\makebox[0pt][l]{\textsuperscript{#2}}\hspace{18pt}}
\newcommand{\benchmissing}[1]{\makebox[34pt][c]{{\fontsize{7}{8}\selectfont\color[HTML]{78818A}#1}}}
\begin{tabular*}{\textwidth}{@{\extracolsep{\fill}}lc*{9}{c}@{}}
\toprule
\textbf{Method} & \textbf{View} & \textbf{OP3} & \textbf{SciPlex} & \textbf{VCPI1} & \textbf{VCPI2} & \textbf{Novartis} & \textbf{Tahoe} & \textbf{P1} & \textbf{P2} & \textbf{Macro} \\
\midrule
\addlinespace[2pt]
\multirow{2}{*}{{\fontsize{9}{10.6}\selectfont Global Mean}} & {\setlength{\fboxsep}{1pt}\colorbox{BenchFullBg}{\makebox[16pt][c]{{\fontsize{6.6}{8}\selectfont\color{BenchFull}Full}}}} & {\color{BenchFull}\benchstat{6.9}{1.0}{}} & {\color{BenchFull}\benchstat{21.2}{0.6}{}} & {\color{BenchFull}\benchstat{20.5}{0.3}{}} & {\color{BenchFull}\benchstat{18.0}{0.1}{}} & {\color{BenchFull}\benchstat{32.2}{0.7}{}} & {\color{BenchFull}\benchstat{20.2}{0.9}{}} & \multirow{2}{*}{\benchstat{8.7}{0.0}{}} & \multirow{2}{*}{\benchstat{4.1}{0.1}{}} & \multirow{2}{*}{\benchstat{16.4}{0.2}{}} \\
 & {\setlength{\fboxsep}{1pt}\colorbox{BenchHVGBg}{\makebox[16pt][c]{{\fontsize{6.6}{8}\selectfont\color{BenchHVG}HVG}}}} & {\color{BenchHVG}\benchstat{7.3}{1.1}{}} & {\color{BenchHVG}\benchstat{14.3}{0.8}{}} & {\color{BenchHVG}\benchstat{15.7}{0.3}{}} & {\color{BenchHVG}\benchstat{13.1}{0.1}{}} & {\color{BenchHVG}\benchstat{27.5}{0.8}{}} & {\color{BenchHVG}\benchstat{20.4}{1.2}{}} &  &  &  \\
\addlinespace[1.5pt]
\multirow{2}{*}{{\fontsize{9}{10.6}\selectfont Linear}} & {\setlength{\fboxsep}{1pt}\colorbox{BenchFullBg}{\makebox[16pt][c]{{\fontsize{6.6}{8}\selectfont\color{BenchFull}Full}}}} & {\color{BenchFull}\benchstat{27.8}{2.3}{}} & {\color{BenchFull}\benchstat{45.9}{1.5}{}} & {\color{BenchFull}\benchstat{14.3}{0.1}{}} & {\color{BenchFull}\benchstat{6.8}{0.4}{}} & {\color{BenchFull}\benchstat{26.9}{0.8}{}} & {\color{BenchFull}\benchstat{35.6}{2.7}{}} & \multirow{2}{*}{\benchstat{15.6}{0.3}{}} & \multirow{2}{*}{\benchstat{2.9}{0.3}{}} & \multirow{2}{*}{\benchstat{23.3}{0.3}{}} \\
 & {\setlength{\fboxsep}{1pt}\colorbox{BenchHVGBg}{\makebox[16pt][c]{{\fontsize{6.6}{8}\selectfont\color{BenchHVG}HVG}}}} & {\color{BenchHVG}\benchstat{30.0}{3.5}{}} & {\color{BenchHVG}\benchstat{43.3}{1.4}{}} & {\color{BenchHVG}\benchstat{11.0}{0.1}{}} & {\color{BenchHVG}\benchstat{5.1}{0.6}{}} & {\color{BenchHVG}\benchstat{23.8}{0.9}{}} & {\color{BenchHVG}\benchstat{36.8}{2.8}{}} &  &  &  \\
\addlinespace[3pt]
\multirow{2}{*}{{\fontsize{9}{10.6}\selectfont Ridge}} & {\setlength{\fboxsep}{1pt}\colorbox{BenchFullBg}{\makebox[16pt][c]{{\fontsize{6.6}{8}\selectfont\color{BenchFull}Full}}}} & {\color{BenchFull}\benchstat{37.5}{1.3}{}} & {\color{BenchFull}\benchstat{50.9}{0.5}{}} & {\color{BenchFull}\benchstat{20.8}{0.4}{}} & {\color{BenchFull}\benchstat{17.9}{0.3}{}} & {\color{BenchFull}\benchstat{33.0}{0.8}{}} & {\color{BenchFull}\benchstat{41.6}{1.1}{}} & \multirow{2}{*}{\benchstat{15.6}{0.3}{}} & \multirow{2}{*}{\benchstat{6.3}{0.1}{}} & \multirow{2}{*}{\benchstat{29.5}{0.3}{}} \\
 & {\setlength{\fboxsep}{1pt}\colorbox{BenchHVGBg}{\makebox[16pt][c]{{\fontsize{6.6}{8}\selectfont\color{BenchHVG}HVG}}}} & {\color{BenchHVG}\benchstat{40.5}{2.0}{}} & {\color{BenchHVG}\benchstat{48.7}{0.6}{}} & {\color{BenchHVG}\benchstat{16.3}{0.4}{}} & {\color{BenchHVG}\benchstat{13.5}{0.2}{}} & {\color{BenchHVG}\benchstat{28.5}{0.9}{}} & {\color{BenchHVG}\benchstat{42.5}{0.9}{}} &  &  &  \\
\addlinespace[1.5pt]
\multirow{2}{*}{{\fontsize{9}{10.6}\selectfont chemCPA}} & {\setlength{\fboxsep}{1pt}\colorbox{BenchFullBg}{\makebox[16pt][c]{{\fontsize{6.6}{8}\selectfont\color{BenchFull}Full}}}} & {\color{BenchFull}\benchstat{29.3}{1.1}{}} & {\color{BenchFull}\benchstat{43.5}{0.6}{}} & {\color{BenchFull}\benchstat{9.7}{0.4}{}} & {\color{BenchFull}\benchstat{9.1}{0.2}{}} & {\color{BenchFull}\benchstat{24.6}{3.0}{}} & {\color{BenchFull}\benchstat{34.3}{2.0}{}} & \multirow{2}{*}{\benchstat{14.3}{0.2}{}} & \multirow{2}{*}{\benchstat{6.0}{0.6}{}} & \multirow{2}{*}{\benchstat{24.5}{0.2}{}} \\
 & {\setlength{\fboxsep}{1pt}\colorbox{BenchHVGBg}{\makebox[16pt][c]{{\fontsize{6.6}{8}\selectfont\color{BenchHVG}HVG}}}} & {\color{BenchHVG}\benchstat{37.0}{0.7}{}} & {\color{BenchHVG}\benchstat{44.7}{0.9}{}} & {\color{BenchHVG}\benchstat{15.6}{0.3}{}} & {\color{BenchHVG}\benchstat{12.9}{0.2}{}} & {\color{BenchHVG}\benchstat{22.1}{0.3}{}} & {\color{BenchHVG}\benchstat{39.3}{1.7}{}} &  &  &  \\
\addlinespace[1.5pt]
\multirow{2}{*}{{\fontsize{9}{10.6}\selectfont PRnet}} & {\setlength{\fboxsep}{1pt}\colorbox{BenchFullBg}{\makebox[16pt][c]{{\fontsize{6.6}{8}\selectfont\color{BenchFull}Full}}}} & {\color{BenchFull}\benchstat{24.7}{7.0}{}} & {\color{BenchFull}\benchstat{47.1}{1.6}{}} & {\color{BenchFull}\benchstat{10.8}{6.4}{}} & {\color{BenchFull}\benchstat{14.6}{0.4}{}} & {\color{BenchFull}\benchstat{30.4}{0.9}{}} & {\color{BenchFull}\benchstat{32.0}{1.7}{}} & \multirow{2}{*}{\benchstat{15.5}{0.2}{}} & \multirow{2}{*}{\benchstat{5.1}{0.2}{}} & \multirow{2}{*}{\benchstat{24.1}{0.6}{}} \\
 & {\setlength{\fboxsep}{1pt}\colorbox{BenchHVGBg}{\makebox[16pt][c]{{\fontsize{6.6}{8}\selectfont\color{BenchHVG}HVG}}}} & {\color{BenchHVG}\benchstat{28.1}{4.9}{}} & {\color{BenchHVG}\benchstat{41.6}{1.9}{}} & {\color{BenchHVG}\benchstat{14.3}{0.5}{}} & {\color{BenchHVG}\benchstat{12.2}{0.1}{}} & {\color{BenchHVG}\benchstat{26.6}{1.3}{}} & {\color{BenchHVG}\benchstat{34.2}{1.3}{}} &  &  &  \\
\addlinespace[1.5pt]
\multirow{2}{*}{{\fontsize{9}{10.6}\selectfont biolord}} & {\setlength{\fboxsep}{1pt}\colorbox{BenchFullBg}{\makebox[16pt][c]{{\fontsize{6.6}{8}\selectfont\color{BenchFull}Full}}}} & {\color{BenchFull}\benchstat{35.4}{3.0}{}} & {\color{BenchFull}\benchstat{49.1}{1.2}{}} & {\color{BenchFull}\benchstat{19.3}{0.2}{}} & {\color{BenchFull}\benchstat{17.0}{0.7}{}} & {\color{BenchFull}\benchstat{31.2}{1.0}{}} & {\color{BenchFull}\benchstat{41.5}{1.2}{}} & \multirow{2}{*}{\benchstat{19.2}{0.8}{}} & \multirow{2}{*}{\benchstat{6.2}{0.3}{}} & \multirow{2}{*}{\benchstat{28.6}{1.0}{}} \\
 & {\setlength{\fboxsep}{1pt}\colorbox{BenchHVGBg}{\makebox[16pt][c]{{\fontsize{6.6}{8}\selectfont\color{BenchHVG}HVG}}}} & {\color{BenchHVG}\benchstat{33.3}{9.3}{}} & {\color{BenchHVG}\benchstat{48.5}{3.0}{}} & {\color{BenchHVG}\benchstat{15.9}{0.1}{}} & {\color{BenchHVG}\benchstat{12.9}{0.4}{}} & {\color{BenchHVG}\benchstat{26.9}{0.8}{}} & {\color{BenchHVG}\benchstat{43.6}{1.0}{}} &  &  &  \\
\addlinespace[1.5pt]
\multirow{2}{*}{{\fontsize{9}{10.6}\selectfont PerturbNet}} & {\setlength{\fboxsep}{1pt}\colorbox{BenchFullBg}{\makebox[16pt][c]{{\fontsize{6.6}{8}\selectfont\color{BenchFull}Full}}}} & {\color{BenchFull}\benchstat{30.7}{3.3}{}} & {\color{BenchFull}\benchstat{48.2}{2.0}{}} & {\color{BenchFull}\benchstat{19.8}{0.6}{}} & {\color{BenchFull}\benchstat{17.6}{0.6}{}} & {\color{BenchFull}\benchstat{30.3}{1.1}{}} & {\color{BenchFull}\benchstat{20.4}{0.6}{}} & \multirow{2}{*}{\benchstat{11.4}{4.2}{}} & \multirow{2}{*}{\benchstat{4.9}{0.2}{}} & \multirow{2}{*}{\benchstat{24.0}{0.5}{}} \\
 & {\setlength{\fboxsep}{1pt}\colorbox{BenchHVGBg}{\makebox[16pt][c]{{\fontsize{6.6}{8}\selectfont\color{BenchHVG}HVG}}}} & {\color{BenchHVG}\benchstat{33.6}{1.2}{}} & {\color{BenchHVG}\benchstat{44.1}{2.0}{}} & {\color{BenchHVG}\benchstat{15.2}{0.6}{}} & {\color{BenchHVG}\benchstat{12.7}{0.4}{}} & {\color{BenchHVG}\benchstat{25.6}{0.8}{}} & {\color{BenchHVG}\benchstat{21.5}{3.9}{}} &  &  &  \\
\addlinespace[1.5pt]
\multirow{2}{*}{{\fontsize{9}{10.6}\selectfont CellFlow}} & {\setlength{\fboxsep}{1pt}\colorbox{BenchFullBg}{\makebox[16pt][c]{{\fontsize{6.6}{8}\selectfont\color{BenchFull}Full}}}} & {\color{BenchFull}\benchstat{36.9}{3.9}{}} & {\color{BenchFull}\benchstat{\underline{51.3}}{0.6}{}} & {\color{BenchFull}\benchstat{21.0}{0.4}{}} & {\color{BenchFull}\benchstat{18.3}{0.4}{}} & {\color{BenchFull}\benchstat{31.6}{0.5}{}} & {\color{BenchFull}\benchstat{43.7}{1.4}{}} & \multirow{2}{*}{\benchmissing{OOT}} & \multirow{2}{*}{\benchmissing{OOT}} & \multirow{2}{*}{\benchstat{\textcolor[HTML]{78818A}{32.9}}{1.1}{\textasteriskcentered}} \\
 & {\setlength{\fboxsep}{1pt}\colorbox{BenchHVGBg}{\makebox[16pt][c]{{\fontsize{6.6}{8}\selectfont\color{BenchHVG}HVG}}}} & {\color{BenchHVG}\benchstat{\underline{40.6}}{6.9}{}} & {\color{BenchHVG}\benchstat{\textbf{49.5}}{0.4}{}} & {\color{BenchHVG}\benchstat{16.2}{0.4}{}} & {\color{BenchHVG}\benchstat{13.4}{0.4}{}} & {\color{BenchHVG}\benchstat{26.9}{0.5}{}} & {\color{BenchHVG}\benchstat{44.9}{1.5}{}} &  &  &  \\
\addlinespace[1.5pt]
\multirow{2}{*}{{\fontsize{9}{10.6}\selectfont CMonge}} & {\setlength{\fboxsep}{1pt}\colorbox{BenchFullBg}{\makebox[16pt][c]{{\fontsize{6.6}{8}\selectfont\color{BenchFull}Full}}}} & {\color{BenchFull}\benchstat{31.8}{4.2}{}} & {\color{BenchFull}\benchstat{\textbf{52.3}}{0.1}{}} & {\color{BenchFull}\benchstat{20.3}{0.2}{}} & {\color{BenchFull}\benchstat{17.9}{0.3}{}} & {\color{BenchFull}\benchstat{31.2}{1.3}{}} & {\color{BenchFull}\benchstat{38.5}{0.4}{}} & \multirow{2}{*}{\benchstat{13.0}{0.3}{}} & \multirow{2}{*}{\benchstat{5.7}{0.2}{}} & \multirow{2}{*}{\benchstat{28.4}{0.7}{}} \\
 & {\setlength{\fboxsep}{1pt}\colorbox{BenchHVGBg}{\makebox[16pt][c]{{\fontsize{6.6}{8}\selectfont\color{BenchHVG}HVG}}}} & {\color{BenchHVG}\benchstat{37.7}{6.0}{}} & {\color{BenchHVG}\benchstat{\underline{49.2}}{0.5}{}} & {\color{BenchHVG}\benchstat{15.6}{0.5}{}} & {\color{BenchHVG}\benchstat{13.2}{0.1}{}} & {\color{BenchHVG}\benchstat{27.0}{0.3}{}} & {\color{BenchHVG}\benchstat{43.9}{0.6}{}} &  &  &  \\
\addlinespace[1.5pt]
\multirow{2}{*}{{\fontsize{9}{10.6}\selectfont Prophet}} & {\setlength{\fboxsep}{1pt}\colorbox{BenchFullBg}{\makebox[16pt][c]{{\fontsize{6.6}{8}\selectfont\color{BenchFull}Full}}}} & {\color{BenchFull}\benchstat{34.9}{2.4}{}} & {\color{BenchFull}\benchstat{48.3}{0.9}{}} & {\color{BenchFull}\benchstat{18.3}{0.7}{}} & {\color{BenchFull}\benchstat{17.6}{0.7}{}} & {\color{BenchFull}\benchstat{29.9}{0.8}{}} & {\color{BenchFull}\benchstat{23.1}{1.1}{}} & \multirow{2}{*}{\benchstat{14.1}{0.1}{}} & \multirow{2}{*}{\benchstat{5.5}{0.3}{}} & \multirow{2}{*}{\benchstat{26.4}{0.6}{}} \\
 & {\setlength{\fboxsep}{1pt}\colorbox{BenchHVGBg}{\makebox[16pt][c]{{\fontsize{6.6}{8}\selectfont\color{BenchHVG}HVG}}}} & {\color{BenchHVG}\benchstat{39.2}{1.7}{}} & {\color{BenchHVG}\benchstat{47.8}{1.0}{}} & {\color{BenchHVG}\benchstat{15.5}{0.3}{}} & {\color{BenchHVG}\benchstat{12.9}{0.0}{}} & {\color{BenchHVG}\benchstat{26.1}{1.3}{}} & {\color{BenchHVG}\benchstat{36.6}{1.5}{}} &  &  &  \\
\addlinespace[1.5pt]
\multirow{2}{*}{{\fontsize{9}{10.6}\selectfont PrePR-CT}} & {\setlength{\fboxsep}{1pt}\colorbox{BenchFullBg}{\makebox[16pt][c]{{\fontsize{6.6}{8}\selectfont\color{BenchFull}Full}}}} & {\color{BenchFull}\benchstat{34.0}{2.7}{}} & {\color{BenchFull}\benchstat{26.1}{8.7}{}} & {\color{BenchFull}\benchstat{20.1}{0.3}{}} & {\color{BenchFull}\benchstat{17.7}{0.1}{}} & {\color{BenchFull}\benchstat{31.8}{0.9}{}} & {\color{BenchFull}\benchstat{20.3}{0.9}{}} & \multirow{2}{*}{\benchstat{9.0}{0.6}{}} & \multirow{2}{*}{\benchstat{4.3}{0.3}{}} & \multirow{2}{*}{\benchstat{25.1}{0.7}{}} \\
 & {\setlength{\fboxsep}{1pt}\colorbox{BenchHVGBg}{\makebox[16pt][c]{{\fontsize{6.6}{8}\selectfont\color{BenchHVG}HVG}}}} & {\color{BenchHVG}\benchstat{39.7}{2.6}{}} & {\color{BenchHVG}\benchstat{48.9}{0.6}{}} & {\color{BenchHVG}\benchstat{15.1}{0.3}{}} & {\color{BenchHVG}\benchstat{12.5}{0.4}{}} & {\color{BenchHVG}\benchstat{28.7}{1.5}{}} & {\color{BenchHVG}\benchstat{43.9}{1.9}{}} &  &  &  \\
\addlinespace[1.5pt]
\multirow{2}{*}{{\fontsize{9}{10.6}\selectfont XPert}} & {\setlength{\fboxsep}{1pt}\colorbox{BenchFullBg}{\makebox[16pt][c]{{\fontsize{6.6}{8}\selectfont\color{BenchFull}Full}}}} & {\color{BenchFull}\benchstat{\underline{38.2}}{1.4}{}} & {\color{BenchFull}\benchmissing{OOM}} & {\color{BenchFull}\benchmissing{OOM}} & {\color{BenchFull}\benchmissing{OOM}} & {\color{BenchFull}\benchmissing{OOM}} & {\color{BenchFull}\benchmissing{OOM}} & \multirow{2}{*}{\benchstat{\underline{29.6}}{0.2}{}} & \multirow{2}{*}{\benchstat{\underline{15.6}}{1.8}{}} & \multirow{2}{*}{\benchstat{\textcolor[HTML]{78818A}{35.4}}{0.3}{\textasteriskcentered}} \\
 & {\setlength{\fboxsep}{1pt}\colorbox{BenchHVGBg}{\makebox[16pt][c]{{\fontsize{6.6}{8}\selectfont\color{BenchHVG}HVG}}}} & {\color{BenchHVG}\benchstat{39.8}{3.4}{}} & {\color{BenchHVG}\benchstat{46.3}{3.4}{}} & {\color{BenchHVG}\benchstat{20.1}{20.3}{}} & {\color{BenchHVG}\benchstat{21.5}{19.9}{}} & {\color{BenchHVG}\benchstat{\textbf{56.5}}{0.7}{}} & {\color{BenchHVG}\benchstat{50.6}{1.4}{}} &  &  &  \\
\addlinespace[1.5pt]
\multirow{2}{*}{{\fontsize{9}{10.6}\selectfont State}} & {\setlength{\fboxsep}{1pt}\colorbox{BenchFullBg}{\makebox[16pt][c]{{\fontsize{6.6}{8}\selectfont\color{BenchFull}Full}}}} & {\color{BenchFull}\benchstat{27.9}{2.2}{}} & {\color{BenchFull}\benchstat{42.4}{1.4}{}} & {\color{BenchFull}\benchstat{\underline{25.2}}{1.0}{}} & {\color{BenchFull}\benchstat{\underline{25.7}}{1.5}{}} & {\color{BenchFull}\benchstat{\underline{47.6}}{0.4}{}} & {\color{BenchFull}\benchstat{\underline{46.8}}{0.5}{}} & \multirow{2}{*}{\benchstat{27.2}{0.2}{}} & \multirow{2}{*}{\benchstat{11.7}{1.4}{}} & \multirow{2}{*}{\benchstat{\underline{33.2}}{0.3}{}} \\
 & {\setlength{\fboxsep}{1pt}\colorbox{BenchHVGBg}{\makebox[16pt][c]{{\fontsize{6.6}{8}\selectfont\color{BenchHVG}HVG}}}} & {\color{BenchHVG}\benchstat{30.7}{3.7}{}} & {\color{BenchHVG}\benchstat{37.2}{0.8}{}} & {\color{BenchHVG}\benchstat{\underline{21.8}}{0.5}{}} & {\color{BenchHVG}\benchstat{\underline{21.6}}{0.8}{}} & {\color{BenchHVG}\benchstat{48.8}{0.6}{}} & {\color{BenchHVG}\benchstat{\underline{50.9}}{1.7}{}} &  &  &  \\
\cmidrule(lr){1-11}
\multirow{2}{*}{{\fontsize{9}{10.6}\selectfont \textbf{Bison}}} & {\setlength{\fboxsep}{1pt}\colorbox{BenchFullBg}{\makebox[16pt][c]{{\fontsize{6.6}{8}\selectfont\color{BenchFull}Full}}}} & {\color{BenchFull}\benchstat{\textbf{39.9}}{1.9}{}} & {\color{BenchFull}\benchstat{45.0}{1.0}{}} & {\color{BenchFull}\benchstat{\textbf{32.5}}{0.5}{}} & {\color{BenchFull}\benchstat{\textbf{32.5}}{0.3}{}} & {\color{BenchFull}\benchstat{\textbf{49.2}}{1.0}{}} & {\color{BenchFull}\benchstat{\textbf{52.0}}{0.3}{}} & \multirow{2}{*}{\benchstat{\textbf{34.3}}{1.0}{}} & \multirow{2}{*}{\benchstat{\textbf{20.7}}{0.7}{}} & \multirow{2}{*}{\benchstat{\textbf{41.1}}{0.6}{}} \\
 & {\setlength{\fboxsep}{1pt}\colorbox{BenchHVGBg}{\makebox[16pt][c]{{\fontsize{6.6}{8}\selectfont\color{BenchHVG}HVG}}}} & {\color{BenchHVG}\benchstat{\textbf{44.2}}{2.8}{}} & {\color{BenchHVG}\benchstat{47.1}{1.6}{}} & {\color{BenchHVG}\benchstat{\textbf{32.7}}{1.3}{}} & {\color{BenchHVG}\benchstat{\textbf{33.3}}{0.9}{}} & {\color{BenchHVG}\benchstat{\underline{53.9}}{1.0}{}} & {\color{BenchHVG}\benchstat{\textbf{58.0}}{0.4}{}} &  &  &  \\
\bottomrule
\end{tabular*}
\endgroup
\end{table}

\begin{table}[H]
\centering
\caption{\textbf{Drug-contrast prediction on globally unseen compounds.} Within-plate drug-contrast Pearson $\times100$, mean (sample SD) across three TEST splits. Full/HVG rows are blue/orange and ranked separately; P1/P2 denote L1000 Phases 1/2. Bold and underline mark the best and second-best displayed means, respectively.}
\label{tab:benchmark_drug_contrast}
\begingroup
\definecolor{BenchFull}{HTML}{2E6185}
\definecolor{BenchHVG}{HTML}{9C5D32}
\definecolor{BenchFullBg}{HTML}{EDF3F8}
\definecolor{BenchHVGBg}{HTML}{F8F0E9}
\fontsize{8}{9.6}\selectfont
\setlength{\tabcolsep}{1.0pt}
\renewcommand{\arraystretch}{1.02}
\newcommand{\benchstat}[3]{\makebox[16pt][r]{#1}\hspace{1pt}\makebox[15pt][r]{{\fontsize{5.8}{7}\selectfont\color{black!62}(#2)}}\makebox[2pt][l]{\textsuperscript{#3}}}
\newcommand{\benchsingle}[2]{\makebox[16pt][r]{#1}\makebox[0pt][l]{\textsuperscript{#2}}\hspace{18pt}}
\newcommand{\benchmissing}[1]{\makebox[34pt][c]{{\fontsize{7}{8}\selectfont\color[HTML]{78818A}#1}}}
\begin{tabular*}{\textwidth}{@{\extracolsep{\fill}}lc*{9}{c}@{}}
\toprule
\textbf{Method} & \textbf{View} & \textbf{OP3} & \textbf{SciPlex} & \textbf{VCPI1} & \textbf{VCPI2} & \textbf{Novartis} & \textbf{Tahoe} & \textbf{P1} & \textbf{P2} & \textbf{Macro} \\
\midrule
\addlinespace[2pt]
\multirow{2}{*}{{\fontsize{9}{10.6}\selectfont Global Mean}} & {\setlength{\fboxsep}{1pt}\colorbox{BenchFullBg}{\makebox[16pt][c]{{\fontsize{6.6}{8}\selectfont\color{BenchFull}Full}}}} & {\color{BenchFull}\benchmissing{NA}} & {\color{BenchFull}\benchmissing{NA}} & {\color{BenchFull}\benchmissing{NA}} & {\color{BenchFull}\benchmissing{NA}} & {\color{BenchFull}\benchmissing{NA}} & {\color{BenchFull}\benchmissing{NA}} & \multirow{2}{*}{\benchmissing{NA}} & \multirow{2}{*}{\benchmissing{NA}} & \multirow{2}{*}{\benchmissing{NA}} \\
 & {\setlength{\fboxsep}{1pt}\colorbox{BenchHVGBg}{\makebox[16pt][c]{{\fontsize{6.6}{8}\selectfont\color{BenchHVG}HVG}}}} & {\color{BenchHVG}\benchmissing{NA}} & {\color{BenchHVG}\benchmissing{NA}} & {\color{BenchHVG}\benchmissing{NA}} & {\color{BenchHVG}\benchmissing{NA}} & {\color{BenchHVG}\benchmissing{NA}} & {\color{BenchHVG}\benchmissing{NA}} &  &  &  \\
\addlinespace[1.5pt]
\multirow{2}{*}{{\fontsize{9}{10.6}\selectfont Linear}} & {\setlength{\fboxsep}{1pt}\colorbox{BenchFullBg}{\makebox[16pt][c]{{\fontsize{6.6}{8}\selectfont\color{BenchFull}Full}}}} & {\color{BenchFull}\benchstat{0.9}{1.5}{}} & {\color{BenchFull}\benchstat{4.5}{0.7}{}} & {\color{BenchFull}\benchstat{0.3}{0.3}{}} & {\color{BenchFull}\benchstat{0.1}{0.4}{}} & {\color{BenchFull}\benchstat{1.8}{0.3}{}} & {\color{BenchFull}\benchstat{1.9}{1.5}{}} & \multirow{2}{*}{\benchstat{2.0}{0.4}{}} & \multirow{2}{*}{\benchstat{0.5}{0.4}{}} & \multirow{2}{*}{\benchstat{1.8}{0.1}{}} \\
 & {\setlength{\fboxsep}{1pt}\colorbox{BenchHVGBg}{\makebox[16pt][c]{{\fontsize{6.6}{8}\selectfont\color{BenchHVG}HVG}}}} & {\color{BenchHVG}\benchstat{0.9}{1.5}{}} & {\color{BenchHVG}\benchstat{6.5}{1.2}{}} & {\color{BenchHVG}\benchstat{0.4}{0.3}{}} & {\color{BenchHVG}\benchstat{0.1}{0.5}{}} & {\color{BenchHVG}\benchstat{2.1}{0.3}{}} & {\color{BenchHVG}\benchstat{2.5}{1.9}{}} &  &  &  \\
\addlinespace[3pt]
\multirow{2}{*}{{\fontsize{9}{10.6}\selectfont Ridge}} & {\setlength{\fboxsep}{1pt}\colorbox{BenchFullBg}{\makebox[16pt][c]{{\fontsize{6.6}{8}\selectfont\color{BenchFull}Full}}}} & {\color{BenchFull}\benchstat{3.7}{4.1}{}} & {\color{BenchFull}\benchstat{6.6}{1.7}{}} & {\color{BenchFull}\benchstat{1.3}{0.4}{}} & {\color{BenchFull}\benchstat{1.3}{0.3}{}} & {\color{BenchFull}\benchstat{\textbf{3.6}}{0.3}{}} & {\color{BenchFull}\benchstat{2.0}{0.9}{}} & \multirow{2}{*}{\benchstat{2.2}{0.4}{}} & \multirow{2}{*}{\benchstat{1.9}{0.2}{}} & \multirow{2}{*}{\benchstat{3.3}{0.3}{}} \\
 & {\setlength{\fboxsep}{1pt}\colorbox{BenchHVGBg}{\makebox[16pt][c]{{\fontsize{6.6}{8}\selectfont\color{BenchHVG}HVG}}}} & {\color{BenchHVG}\benchstat{4.6}{4.6}{}} & {\color{BenchHVG}\benchstat{8.9}{2.2}{}} & {\color{BenchHVG}\benchstat{\underline{1.7}}{0.5}{}} & {\color{BenchHVG}\benchstat{\underline{1.6}}{0.3}{}} & {\color{BenchHVG}\benchstat{\underline{4.3}}{0.4}{}} & {\color{BenchHVG}\benchstat{2.3}{1.0}{}} &  &  &  \\
\addlinespace[1.5pt]
\multirow{2}{*}{{\fontsize{9}{10.6}\selectfont chemCPA}} & {\setlength{\fboxsep}{1pt}\colorbox{BenchFullBg}{\makebox[16pt][c]{{\fontsize{6.6}{8}\selectfont\color{BenchFull}Full}}}} & {\color{BenchFull}\benchstat{0.9}{1.2}{}} & {\color{BenchFull}\benchstat{\underline{8.1}}{1.8}{}} & {\color{BenchFull}\benchstat{0.6}{0.5}{}} & {\color{BenchFull}\benchstat{0.4}{0.4}{}} & {\color{BenchFull}\benchstat{0.2}{0.5}{}} & {\color{BenchFull}\benchstat{1.2}{2.8}{}} & \multirow{2}{*}{\benchstat{2.0}{0.8}{}} & \multirow{2}{*}{\benchstat{1.2}{0.5}{}} & \multirow{2}{*}{\benchstat{2.0}{0.5}{}} \\
 & {\setlength{\fboxsep}{1pt}\colorbox{BenchHVGBg}{\makebox[16pt][c]{{\fontsize{6.6}{8}\selectfont\color{BenchHVG}HVG}}}} & {\color{BenchHVG}\benchstat{0.9}{3.4}{}} & {\color{BenchHVG}\benchstat{9.6}{2.8}{}} & {\color{BenchHVG}\benchstat{1.0}{0.8}{}} & {\color{BenchHVG}\benchstat{0.0}{0.7}{}} & {\color{BenchHVG}\benchstat{0.7}{0.7}{}} & {\color{BenchHVG}\benchstat{0.7}{2.1}{}} &  &  &  \\
\addlinespace[1.5pt]
\multirow{2}{*}{{\fontsize{9}{10.6}\selectfont PRnet}} & {\setlength{\fboxsep}{1pt}\colorbox{BenchFullBg}{\makebox[16pt][c]{{\fontsize{6.6}{8}\selectfont\color{BenchFull}Full}}}} & {\color{BenchFull}\benchstat{0.5}{1.7}{}} & {\color{BenchFull}\benchstat{\textbf{8.1}}{2.4}{}} & {\color{BenchFull}\benchstat{0.1}{0.4}{}} & {\color{BenchFull}\benchstat{0.3}{0.4}{}} & {\color{BenchFull}\benchstat{1.1}{0.3}{}} & {\color{BenchFull}\benchstat{0.6}{1.0}{}} & \multirow{2}{*}{\benchstat{1.2}{0.3}{}} & \multirow{2}{*}{\benchstat{0.3}{0.2}{}} & \multirow{2}{*}{\benchstat{1.9}{0.3}{}} \\
 & {\setlength{\fboxsep}{1pt}\colorbox{BenchHVGBg}{\makebox[16pt][c]{{\fontsize{6.6}{8}\selectfont\color{BenchHVG}HVG}}}} & {\color{BenchHVG}\benchstat{-0.4}{2.9}{}} & {\color{BenchHVG}\benchstat{\textbf{11.2}}{3.5}{}} & {\color{BenchHVG}\benchstat{0.4}{0.3}{}} & {\color{BenchHVG}\benchstat{0.6}{0.0}{}} & {\color{BenchHVG}\benchstat{1.4}{0.2}{}} & {\color{BenchHVG}\benchstat{1.1}{0.4}{}} &  &  &  \\
\addlinespace[1.5pt]
\multirow{2}{*}{{\fontsize{9}{10.6}\selectfont biolord}} & {\setlength{\fboxsep}{1pt}\colorbox{BenchFullBg}{\makebox[16pt][c]{{\fontsize{6.6}{8}\selectfont\color{BenchFull}Full}}}} & {\color{BenchFull}\benchstat{-0.4}{1.1}{}} & {\color{BenchFull}\benchstat{4.3}{4.1}{}} & {\color{BenchFull}\benchstat{-0.1}{0.1}{}} & {\color{BenchFull}\benchstat{0.0}{0.0}{}} & {\color{BenchFull}\benchstat{1.0}{0.4}{}} & {\color{BenchFull}\benchstat{0.9}{1.8}{}} & \multirow{2}{*}{\benchstat{1.4}{0.4}{}} & \multirow{2}{*}{\benchstat{0.8}{0.4}{}} & \multirow{2}{*}{\benchstat{1.6}{0.7}{}} \\
 & {\setlength{\fboxsep}{1pt}\colorbox{BenchHVGBg}{\makebox[16pt][c]{{\fontsize{6.6}{8}\selectfont\color{BenchHVG}HVG}}}} & {\color{BenchHVG}\benchstat{-0.2}{0.9}{}} & {\color{BenchHVG}\benchstat{\underline{9.9}}{6.4}{}} & {\color{BenchHVG}\benchstat{0.6}{0.6}{}} & {\color{BenchHVG}\benchstat{0.0}{0.6}{}} & {\color{BenchHVG}\benchstat{1.1}{0.5}{}} & {\color{BenchHVG}\benchstat{2.9}{1.8}{}} &  &  &  \\
\addlinespace[1.5pt]
\multirow{2}{*}{{\fontsize{9}{10.6}\selectfont PerturbNet}} & {\setlength{\fboxsep}{1pt}\colorbox{BenchFullBg}{\makebox[16pt][c]{{\fontsize{6.6}{8}\selectfont\color{BenchFull}Full}}}} & {\color{BenchFull}\benchstat{0.7}{0.9}{}} & {\color{BenchFull}\benchstat{3.2}{1.9}{}} & {\color{BenchFull}\benchstat{0.6}{0.6}{}} & {\color{BenchFull}\benchstat{0.3}{0.5}{}} & {\color{BenchFull}\benchstat{1.3}{0.2}{}} & {\color{BenchFull}\benchstat{0.4}{1.0}{}} & \multirow{2}{*}{\benchstat{0.2}{0.3}{}} & \multirow{2}{*}{\benchstat{0.5}{0.2}{}} & \multirow{2}{*}{\benchstat{1.1}{0.1}{}} \\
 & {\setlength{\fboxsep}{1pt}\colorbox{BenchHVGBg}{\makebox[16pt][c]{{\fontsize{6.6}{8}\selectfont\color{BenchHVG}HVG}}}} & {\color{BenchHVG}\benchstat{0.2}{2.6}{}} & {\color{BenchHVG}\benchstat{4.0}{2.1}{}} & {\color{BenchHVG}\benchstat{0.5}{0.4}{}} & {\color{BenchHVG}\benchstat{0.8}{0.4}{}} & {\color{BenchHVG}\benchstat{2.1}{0.9}{}} & {\color{BenchHVG}\benchstat{0.1}{0.5}{}} &  &  &  \\
\addlinespace[1.5pt]
\multirow{2}{*}{{\fontsize{9}{10.6}\selectfont CellFlow}} & {\setlength{\fboxsep}{1pt}\colorbox{BenchFullBg}{\makebox[16pt][c]{{\fontsize{6.6}{8}\selectfont\color{BenchFull}Full}}}} & {\color{BenchFull}\benchstat{-2.2}{1.2}{}} & {\color{BenchFull}\benchstat{4.2}{2.7}{}} & {\color{BenchFull}\benchstat{0.5}{0.5}{}} & {\color{BenchFull}\benchstat{0.4}{0.3}{}} & {\color{BenchFull}\benchstat{1.3}{0.7}{}} & {\color{BenchFull}\benchstat{0.8}{1.2}{}} & \multirow{2}{*}{\benchmissing{OOT}} & \multirow{2}{*}{\benchmissing{OOT}} & \multirow{2}{*}{\benchstat{\textcolor[HTML]{78818A}{1.4}}{0.1}{\textasteriskcentered}} \\
 & {\setlength{\fboxsep}{1pt}\colorbox{BenchHVGBg}{\makebox[16pt][c]{{\fontsize{6.6}{8}\selectfont\color{BenchHVG}HVG}}}} & {\color{BenchHVG}\benchstat{-1.4}{1.6}{}} & {\color{BenchHVG}\benchstat{9.1}{1.4}{}} & {\color{BenchHVG}\benchstat{0.7}{0.6}{}} & {\color{BenchHVG}\benchstat{0.3}{0.3}{}} & {\color{BenchHVG}\benchstat{1.5}{0.7}{}} & {\color{BenchHVG}\benchstat{1.2}{1.2}{}} &  &  &  \\
\addlinespace[1.5pt]
\multirow{2}{*}{{\fontsize{9}{10.6}\selectfont CMonge}} & {\setlength{\fboxsep}{1pt}\colorbox{BenchFullBg}{\makebox[16pt][c]{{\fontsize{6.6}{8}\selectfont\color{BenchFull}Full}}}} & {\color{BenchFull}\benchstat{-0.8}{1.2}{}} & {\color{BenchFull}\benchstat{6.3}{2.3}{}} & {\color{BenchFull}\benchstat{0.0}{0.1}{}} & {\color{BenchFull}\benchstat{0.1}{0.2}{}} & {\color{BenchFull}\benchstat{0.5}{0.5}{}} & {\color{BenchFull}\benchstat{0.4}{0.8}{}} & \multirow{2}{*}{\benchstat{0.1}{0.1}{}} & \multirow{2}{*}{\benchstat{0.3}{0.4}{}} & \multirow{2}{*}{\benchstat{0.8}{0.3}{}} \\
 & {\setlength{\fboxsep}{1pt}\colorbox{BenchHVGBg}{\makebox[16pt][c]{{\fontsize{6.6}{8}\selectfont\color{BenchHVG}HVG}}}} & {\color{BenchHVG}\benchstat{-0.5}{1.5}{}} & {\color{BenchHVG}\benchstat{3.6}{2.2}{}} & {\color{BenchHVG}\benchstat{0.0}{0.4}{}} & {\color{BenchHVG}\benchstat{0.2}{0.3}{}} & {\color{BenchHVG}\benchstat{0.6}{0.4}{}} & {\color{BenchHVG}\benchstat{0.3}{0.4}{}} &  &  &  \\
\addlinespace[1.5pt]
\multirow{2}{*}{{\fontsize{9}{10.6}\selectfont Prophet}} & {\setlength{\fboxsep}{1pt}\colorbox{BenchFullBg}{\makebox[16pt][c]{{\fontsize{6.6}{8}\selectfont\color{BenchFull}Full}}}} & {\color{BenchFull}\benchstat{-0.9}{2.0}{}} & {\color{BenchFull}\benchstat{3.0}{0.9}{}} & {\color{BenchFull}\benchstat{0.7}{0.2}{}} & {\color{BenchFull}\benchstat{0.5}{0.3}{}} & {\color{BenchFull}\benchstat{0.6}{0.8}{}} & {\color{BenchFull}\benchstat{2.1}{0.1}{}} & \multirow{2}{*}{\benchstat{0.7}{0.2}{}} & \multirow{2}{*}{\benchstat{0.9}{0.4}{}} & \multirow{2}{*}{\benchstat{1.4}{0.2}{}} \\
 & {\setlength{\fboxsep}{1pt}\colorbox{BenchHVGBg}{\makebox[16pt][c]{{\fontsize{6.6}{8}\selectfont\color{BenchHVG}HVG}}}} & {\color{BenchHVG}\benchstat{2.0}{2.4}{}} & {\color{BenchHVG}\benchstat{5.2}{1.0}{}} & {\color{BenchHVG}\benchstat{0.6}{0.6}{}} & {\color{BenchHVG}\benchstat{0.1}{0.3}{}} & {\color{BenchHVG}\benchstat{1.0}{0.2}{}} & {\color{BenchHVG}\benchstat{2.3}{2.2}{}} &  &  &  \\
\addlinespace[1.5pt]
\multirow{2}{*}{{\fontsize{9}{10.6}\selectfont PrePR-CT}} & {\setlength{\fboxsep}{1pt}\colorbox{BenchFullBg}{\makebox[16pt][c]{{\fontsize{6.6}{8}\selectfont\color{BenchFull}Full}}}} & {\color{BenchFull}\benchstat{-0.3}{1.2}{}} & {\color{BenchFull}\benchstat{0.0}{0.0}{}} & {\color{BenchFull}\benchstat{0.1}{0.2}{}} & {\color{BenchFull}\benchstat{0.0}{0.0}{}} & {\color{BenchFull}\benchstat{-0.1}{0.2}{}} & {\color{BenchFull}\benchstat{0.0}{0.0}{}} & \multirow{2}{*}{\benchstat{-0.1}{0.1}{}} & \multirow{2}{*}{\benchstat{0.3}{0.4}{}} & \multirow{2}{*}{\benchstat{0.8}{0.5}{}} \\
 & {\setlength{\fboxsep}{1pt}\colorbox{BenchHVGBg}{\makebox[16pt][c]{{\fontsize{6.6}{8}\selectfont\color{BenchHVG}HVG}}}} & {\color{BenchHVG}\benchstat{-0.4}{3.9}{}} & {\color{BenchHVG}\benchstat{9.9}{3.7}{}} & {\color{BenchHVG}\benchstat{-0.3}{0.9}{}} & {\color{BenchHVG}\benchstat{0.2}{0.5}{}} & {\color{BenchHVG}\benchstat{1.7}{0.5}{}} & {\color{BenchHVG}\benchstat{0.8}{0.2}{}} &  &  &  \\
\addlinespace[1.5pt]
\multirow{2}{*}{{\fontsize{9}{10.6}\selectfont XPert}} & {\setlength{\fboxsep}{1pt}\colorbox{BenchFullBg}{\makebox[16pt][c]{{\fontsize{6.6}{8}\selectfont\color{BenchFull}Full}}}} & {\color{BenchFull}\benchstat{-1.3}{0.4}{}} & {\color{BenchFull}\benchmissing{OOM}} & {\color{BenchFull}\benchmissing{OOM}} & {\color{BenchFull}\benchmissing{OOM}} & {\color{BenchFull}\benchmissing{OOM}} & {\color{BenchFull}\benchmissing{OOM}} & \multirow{2}{*}{\benchstat{\underline{4.9}}{1.0}{}} & \multirow{2}{*}{\benchstat{\underline{4.1}}{0.2}{}} & \multirow{2}{*}{\benchstat{\textcolor[HTML]{78818A}{2.5}}{0.3}{\textasteriskcentered}} \\
 & {\setlength{\fboxsep}{1pt}\colorbox{BenchHVGBg}{\makebox[16pt][c]{{\fontsize{6.6}{8}\selectfont\color{BenchHVG}HVG}}}} & {\color{BenchHVG}\benchstat{0.1}{0.6}{}} & {\color{BenchHVG}\benchstat{4.7}{1.1}{}} & {\color{BenchHVG}\benchmissing{NA}} & {\color{BenchHVG}\benchstat{0.9}{1.1}{}} & {\color{BenchHVG}\benchstat{\textbf{5.0}}{0.7}{}} & {\color{BenchHVG}\benchstat{1.7}{1.3}{}} &  &  &  \\
\addlinespace[1.5pt]
\multirow{2}{*}{{\fontsize{9}{10.6}\selectfont State}} & {\setlength{\fboxsep}{1pt}\colorbox{BenchFullBg}{\makebox[16pt][c]{{\fontsize{6.6}{8}\selectfont\color{BenchFull}Full}}}} & {\color{BenchFull}\benchstat{\underline{5.0}}{1.9}{}} & {\color{BenchFull}\benchstat{5.1}{1.3}{}} & {\color{BenchFull}\benchstat{\underline{1.4}}{0.5}{}} & {\color{BenchFull}\benchstat{\underline{1.6}}{0.1}{}} & {\color{BenchFull}\benchstat{2.9}{0.3}{}} & {\color{BenchFull}\benchstat{\textbf{3.2}}{0.9}{}} & \multirow{2}{*}{\benchstat{3.3}{0.6}{}} & \multirow{2}{*}{\benchstat{2.7}{0.1}{}} & \multirow{2}{*}{\benchstat{\underline{3.4}}{0.1}{}} \\
 & {\setlength{\fboxsep}{1pt}\colorbox{BenchHVGBg}{\makebox[16pt][c]{{\fontsize{6.6}{8}\selectfont\color{BenchHVG}HVG}}}} & {\color{BenchHVG}\benchstat{\underline{4.7}}{2.0}{}} & {\color{BenchHVG}\benchstat{7.3}{1.5}{}} & {\color{BenchHVG}\benchstat{1.4}{0.0}{}} & {\color{BenchHVG}\benchstat{1.1}{0.6}{}} & {\color{BenchHVG}\benchstat{3.7}{0.3}{}} & {\color{BenchHVG}\benchstat{\textbf{4.4}}{0.7}{}} &  &  &  \\
\cmidrule(lr){1-11}
\multirow{2}{*}{{\fontsize{9}{10.6}\selectfont \textbf{Bison}}} & {\setlength{\fboxsep}{1pt}\colorbox{BenchFullBg}{\makebox[16pt][c]{{\fontsize{6.6}{8}\selectfont\color{BenchFull}Full}}}} & {\color{BenchFull}\benchstat{\textbf{8.3}}{1.9}{}} & {\color{BenchFull}\benchstat{6.2}{2.1}{}} & {\color{BenchFull}\benchstat{\textbf{2.6}}{0.3}{}} & {\color{BenchFull}\benchstat{\textbf{1.7}}{0.5}{}} & {\color{BenchFull}\benchstat{\underline{3.5}}{1.0}{}} & {\color{BenchFull}\benchstat{\underline{2.6}}{2.0}{}} & \multirow{2}{*}{\benchstat{\textbf{5.0}}{1.1}{}} & \multirow{2}{*}{\benchstat{\textbf{4.2}}{0.7}{}} & \multirow{2}{*}{\benchstat{\textbf{4.8}}{0.4}{}} \\
 & {\setlength{\fboxsep}{1pt}\colorbox{BenchHVGBg}{\makebox[16pt][c]{{\fontsize{6.6}{8}\selectfont\color{BenchHVG}HVG}}}} & {\color{BenchHVG}\benchstat{\textbf{9.9}}{1.9}{}} & {\color{BenchHVG}\benchstat{9.5}{2.9}{}} & {\color{BenchHVG}\benchstat{\textbf{3.4}}{0.3}{}} & {\color{BenchHVG}\benchstat{\textbf{2.3}}{0.6}{}} & {\color{BenchHVG}\benchstat{4.2}{1.2}{}} & {\color{BenchHVG}\benchstat{\underline{3.4}}{2.7}{}} &  &  &  \\
\bottomrule
\end{tabular*}
\endgroup
\end{table}

\subsection{Joint and Dataset-Specific Training}
\label{app:joint_solo_details}

The overall-response columns of Table~\ref{tab:joint_solo} vary the context-response and molecular-effect DLM training regimes independently on the first compound split.
The JOINT checkpoints are update 80,000 for the context-response branch and 40,000 for the molecular-effect branch.
These are the first split's main-benchmark checkpoints, evaluated here before calibration with canonical recombined-replicate scoring.
The six count-based SOLO context models use their VAL-selected checkpoints; the L1000 context models select updates 32,002 and 22,182 for P1 and P2 by complete soft-decoded VAL evaluation.
Molecular SOLO models select by VAL drug contrast within budgets of 1,500 (OP3), 2,000 (SciPlex), 2,000 (VCPI1), 3,000 (VCPI2), 6,000 (Novartis), 2,000 (Tahoe), 16,000 (L1000 P1) and 5,000 (L1000 P2) updates.
The comparison therefore uses a shared representation and matched per-dataset training budgets, with each SOLO run allocated the same training quota as its dataset receives under joint training, proportional to its share of profiles in the joint training mixture.

Drug contrast compares molecular SOLO and JOINT training with the context-response branch fixed to JOINT.
We average Full/HVG evaluation routes within each dataset; both L1000 routes use the landmark panel.
Both Table~\ref{tab:joint_solo} and Figure~\ref{fig:joint_solo_scatter} report these dataset means, with equal dataset weights in the Macro: 0.03974 for SOLO and 0.05063 for JOINT, a 27.4\% increase.
Overall-response entries retain the common first split.

\subsection{Ablations}
\label{app:ablations}
\label{sec:component_analysis}

\paragraph{Molecular contribution and scale.}\label{sec:molecular_contribution}
Adding the molecular deviation improves uncalibrated canonical overall Pearson from 0.4046 to 0.4123, with a paired increment of $0.0077\pm0.0051$ across three TEST splits (Table~\ref{tab:ablation_summary}a).
Moderate deviation weights improve the shared response, whereas $\alpha=1$ reduces it (panel b).
The fixed $\alpha=0.25$ gives the largest gain among the tested weights, $+0.0091$.
Drug contrast varies little across positive weights because rescaling largely preserves its direction.

\begingroup
\setlength{\textfloatsep}{10pt plus 2pt minus 1pt}
\begin{table}[H]
\centering
\small
\caption{\textbf{Ablations of Bison.}
(a) Uncalibrated canonical scores: mean (sample SD) and paired increments.
(b) Overall-Pearson gain over $\alpha=0$ with both DLMs fixed; $\dagger$: fixed coefficient.
(c) Drug contrast: Context combines context modulation and interaction loss; Contrast is the within-plate loss.
Panels (b,c) weight datasets equally after averaging native views.}
\label{tab:ablation_summary}
\label{tab:response_composition}
\label{tab:alpha_ablation}
\label{tab:component_ablation}
\setlength{\tabcolsep}{5pt}
\renewcommand{\arraystretch}{1.12}
\newcommand{\ablationsd}[1]{\,{\footnotesize\textcolor{black!65}{(#1)}}}
\newcommand{\ablationheading}[2]{\noindent\makebox[\linewidth][l]{\textbf{#1}\hfill{\footnotesize\textcolor{black!65}{#2}}}\par\vspace{3pt}}

\ablationheading{(a) Response composition}{Three TEST splits; 14-view mean}
\begin{tabular*}{\linewidth}{@{\extracolsep{\fill}}lrrr@{}}
\toprule
Predictor & Overall Pearson $\Delta$ & Increment & Drug contrast \\
\midrule
Context-response DLM & 0.4046\ablationsd{0.0033} & -- & -- \\
Bison (two DLMs) & 0.4123\ablationsd{0.0055} & $+0.0077$\ablationsd{0.0051} & 0.0477\ablationsd{0.0041} \\
\bottomrule
\end{tabular*}
\par\vspace{8pt}

\ablationheading{(b) Molecular deviation weight}{VAL}
\begin{tabular*}{\linewidth}{@{\extracolsep{\fill}}lrrrrr@{}}
\toprule
Deviation weight $\alpha$ & 0 & 0.125 & $0.25^{\dagger}$ & 0.5 & 1 \\
\midrule
Overall Pearson gain & 0.0000 & $+0.0075$ & $+0.0091$ & $+0.0022$ & $-0.0278$ \\
\bottomrule
\end{tabular*}
\par\vspace{8pt}

\ablationheading{(c) Molecular-effect components}{VAL}
\begin{tabular*}{\linewidth}{@{\extracolsep{\fill}}ccrrr@{}}
\toprule
Context components & Contrast loss & Eight-dataset mean & OP3 & Tahoe \\
\midrule
-- & -- & 0.0246 & $-0.0223$ & 0.0217 \\
-- & \checkmark & 0.0288 & $-0.0198$ & 0.0537 \\
\checkmark & -- & 0.0358 & 0.0440 & 0.0480 \\
\checkmark & \checkmark & \textbf{0.0369} & 0.0329 & \textbf{0.0546} \\
\bottomrule
\end{tabular*}
\end{table}
\endgroup

\paragraph{Context modeling and contrast supervision.}
Adding context modulation and interaction supervision raises three-split drug contrast from 0.0414 to 0.0477 and native overall Pearson from 0.4082 to 0.4097 before calibration.
The factorial comparison in Table~\ref{tab:ablation_summary}c separates this context-component bundle from within-plate contrast supervision.
Context components raise the mean from 0.0288 to 0.0369 with contrast supervision and from 0.0246 to 0.0358 without it.
They account for most of the aggregate gain, while the within-plate loss adds a smaller increment; OP3 and Tahoe show different responses to the two components.

\paragraph{Evaluation settings.}
Panel (a) uses uncalibrated canonical recombined-replicate scoring over fourteen TEST views and three splits, with the control-coded molecular reference and $\alpha=0.25$.
Panel (b) uses the first split's VAL data, the selected molecular checkpoint at update 40,000 and the same control-coded reference.
Panel (c) compares all four molecular variants at update 35,000 on the third split's VAL data.
Panels (b,c) average views within datasets before weighting datasets equally.
The three-split context-design comparison holds the shared branch, reference convention and composition weight fixed.

\subsection{Calibration Effects}
\label{app:calibration_results}

Table~\ref{tab:postprocessing_tradeoffs} compares the uncalibrated composition, amplitude calibration and detection-aware zeroing from the same branch predictions.
Amplitude calibration increases PDS-L1 from 0.7028 to 0.7778 and PDS-L2 from 0.6909 to 0.7647, with MAE increasing from 0.0537 to 0.0569.
Detection-aware zeroing increases overlap@100 from 0.0137 to 0.1670 and improves gene-effect ranking, while DE direction agreement decreases from 0.7797 to 0.7006.
Overall Pearson changes little across the three outputs.
Each variant uses complete condition sets and the same defined views across all three splits; estimation details are in Appendix~\ref{app:calibration_details}.

\begin{table}[H]
\centering
\caption{\textbf{Effects and tradeoffs of postprocessing.} Three-split means and sample SD on fixed common Cell-Eval views. Values are on their original scales.}
\label{tab:postprocessing_tradeoffs}
\fontsize{8}{10}\selectfont
\setlength{\tabcolsep}{4pt}
\begin{tabular}{lcrrr}
\toprule
Metric & Views & Uncalibrated & + amplitude & + detection zeros \\
\midrule
Pearson $\Delta\uparrow$ & 12 & 0.4321 (0.0055) & 0.4336 (0.0055) & 0.4337 (0.0055) \\
MAE $\Delta\downarrow$ & 12 & 0.0537 (0.0006) & 0.0569 (0.0008) & 0.0568 (0.0008) \\
MSE $\Delta\downarrow$ & 12 & 0.0112 (0.0005) & 0.0120 (0.0005) & 0.0120 (0.0005) \\
PDS-L1 $\uparrow$ & 12 & 0.7028 (0.0030) & 0.7778 (0.0049) & 0.7782 (0.0048) \\
PDS-L2 $\uparrow$ & 12 & 0.6909 (0.0026) & 0.7647 (0.0052) & 0.7647 (0.0052) \\
PDS-cosine $\uparrow$ & 12 & 0.8430 (0.0040) & 0.8458 (0.0038) & 0.8458 (0.0038) \\
Pearson E-distance $\uparrow$ & 12 & 0.3866 (0.0309) & 0.3690 (0.0303) & 0.3690 (0.0303) \\
Clustering agreement $\uparrow$ & 12 & 0.5265 (0.0160) & 0.5270 (0.0165) & 0.5270 (0.0165) \\
DE direction $\uparrow$ & 10 & 0.7804 (0.0080) & 0.7797 (0.0084) & 0.7006 (0.0072) \\
DE-count Spearman $\uparrow$ & 9 & 0.1979 (0.1159) & 0.2010 (0.0537) & 0.3310 (0.0621) \\
DE log-fold-change Spearman $\uparrow$ & 10 & 0.5785 (0.0107) & 0.6025 (0.0117) & 0.6307 (0.0155) \\
DE recall $\uparrow$ & 10 & 0.4929 (0.0272) & 0.5282 (0.0293) & 0.4347 (0.0277) \\
ROC-AUC $\uparrow$ & 10 & 0.7028 (0.0063) & 0.7027 (0.0066) & 0.6807 (0.0060) \\
PR-AUC $\uparrow$ & 10 & 0.2323 (0.0293) & 0.2381 (0.0258) & 0.3852 (0.0227) \\
Overlap@100 $\uparrow$ & 12 & 0.0069 (0.0005) & 0.0137 (0.0005) & 0.1670 (0.0109) \\
Precision@100 $\uparrow$ & 12 & 0.0064 (0.0003) & 0.0133 (0.0006) & 0.1652 (0.0106) \\
\bottomrule
\end{tabular}
\par\vspace{4pt}
\begin{minipage}{\textwidth}\footnotesize
All Bison rows use complete condition sets; undefined DE scores are excluded using the same fixed view set across variants and splits. MAE/MSE equal their delta versions under the matched-control subtraction and are not duplicated. Amplitude calibration improves L1/L2 discrimination but increases magnitude errors; detection-aware zeros improve DE ranking and overlap while lowering DE direction agreement, recall and ROC-AUC relative to amplitude calibration alone. All three variants are computed from the same branch predictions within each split.
\end{minipage}
\end{table}

\subsection{Control-Reference Sensitivity}
\label{app:reference_sensitivity}

Table~\ref{tab:control_reference_sensitivity} scores the same final predictions relative to pooled context--time and matched context--plate--time TRAIN controls, using native condition aggregation in both cases.
Bison has the highest complete-coverage mean under both conventions: 0.5637 with pooled controls and 0.4110 with plate matching.
Thus, the overall ranking persists when the scoring reference changes, while drug contrast separately evaluates compound-dependent differences.

\begin{table}[H]
\centering
\caption{\textbf{Sensitivity to the control reference.} Native Pearson $\Delta$, mean (sample SD) across three TEST splits. Identical stored predictions are scored against pooled context--time and matched context--plate--time controls. Each split averages a fixed common set of official views. Partial means (*) are gray and excluded from full-coverage ranking; bold/underline mark the best/second-best mean. Bison uses the same corrected molecular reference as Table~\ref{tab:benchmark_results}.}
\label{tab:control_reference_sensitivity}
\small
\begin{tabular}{lrrr}
\toprule
Method & Context--time & Plate matched & Views \\
\midrule
Ridge & 0.2320\,{\fontsize{6.5}{8}\selectfont\textcolor{black!62}{(0.0049)}} & 0.2954\,{\fontsize{6.5}{8}\selectfont\textcolor{black!62}{(0.0034)}} & 14/14 \\
chemCPA & 0.1492\,{\fontsize{6.5}{8}\selectfont\textcolor{black!62}{(0.0034)}} & 0.2445\,{\fontsize{6.5}{8}\selectfont\textcolor{black!62}{(0.0019)}} & 14/14 \\
PRnet & 0.1603\,{\fontsize{6.5}{8}\selectfont\textcolor{black!62}{(0.0025)}} & 0.2408\,{\fontsize{6.5}{8}\selectfont\textcolor{black!62}{(0.0062)}} & 14/14 \\
biolord & 0.2029\,{\fontsize{6.5}{8}\selectfont\textcolor{black!62}{(0.0113)}} & 0.2856\,{\fontsize{6.5}{8}\selectfont\textcolor{black!62}{(0.0103)}} & 14/14 \\
PerturbNet & 0.1536\,{\fontsize{6.5}{8}\selectfont\textcolor{black!62}{(0.0127)}} & 0.2400\,{\fontsize{6.5}{8}\selectfont\textcolor{black!62}{(0.0051)}} & 14/14 \\
CellFlow & \textcolor{black!50}{0.2493$^{*}$}\,{\fontsize{6.5}{8}\selectfont\textcolor{black!62}{(0.0199)}} & \textcolor{black!50}{0.3288$^{*}$}\,{\fontsize{6.5}{8}\selectfont\textcolor{black!62}{(0.0110)}} & 12/14 \\
CMonge & 0.1845\,{\fontsize{6.5}{8}\selectfont\textcolor{black!62}{(0.0073)}} & 0.2837\,{\fontsize{6.5}{8}\selectfont\textcolor{black!62}{(0.0073)}} & 14/14 \\
Prophet & 0.1687\,{\fontsize{6.5}{8}\selectfont\textcolor{black!62}{(0.0071)}} & 0.2642\,{\fontsize{6.5}{8}\selectfont\textcolor{black!62}{(0.0064)}} & 14/14 \\
PrePR-CT & 0.1573\,{\fontsize{6.5}{8}\selectfont\textcolor{black!62}{(0.0012)}} & 0.2514\,{\fontsize{6.5}{8}\selectfont\textcolor{black!62}{(0.0072)}} & 14/14 \\
XPert & \textcolor{black!50}{0.5882$^{*}$}\,{\fontsize{6.5}{8}\selectfont\textcolor{black!62}{(0.0032)}} & \textcolor{black!50}{0.3536$^{*}$}\,{\fontsize{6.5}{8}\selectfont\textcolor{black!62}{(0.0028)}} & 9/14 \\
State & \underline{0.5028}\,{\fontsize{6.5}{8}\selectfont\textcolor{black!62}{(0.0042)}} & \underline{0.3325}\,{\fontsize{6.5}{8}\selectfont\textcolor{black!62}{(0.0029)}} & 14/14 \\
Bison & \textbf{0.5637}\,{\fontsize{6.5}{8}\selectfont\textcolor{black!62}{(0.0035)}} & \textbf{0.4110}\,{\fontsize{6.5}{8}\selectfont\textcolor{black!62}{(0.0059)}} & 14/14 \\
\bottomrule
\end{tabular}
\end{table}

\clearpage
\section{Transfer and Pathways}
\label{app:interpretability}

\subsection{Source Contributions and Structural Proximity}
\label{app:source_removal_statistics}

\paragraph{Source-removal design.}
Each source-removal model omits one dataset's molecular-effect DLM updates while preserving the retained updates' order and learning rates.
We compare against the complete joint model at original schedule step 20,000 on the first split's VAL compounds.
The tokenizer and context-response branch remain shared across all eight datasets, so this measures the contribution of a molecular training source within the shared representation.
Scores average native views within each dataset, with a single landmark view for L1000.

\begin{figure}[H]
\centering
\includegraphics[width=0.85\textwidth]{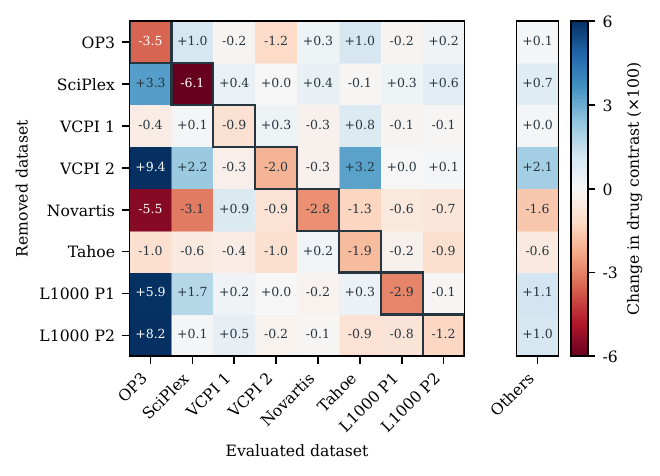}
\caption{\textbf{Dataset-level source-removal effects.} Change in VAL drug contrast ($\times100$) relative to joint training at original schedule step 20,000. Rows identify the removed source and columns the evaluated dataset; outlined diagonal cells remove the target dataset itself. The right column averages each row over the other seven datasets.}
\label{fig:source_removal}
\end{figure}

\paragraph{Source and target contributions.}
Figure~\ref{fig:source_removal} separates each source's effect on its own predictions from its effect on the other datasets.
Removing a dataset lowers its own drug contrast in all eight cases.
Novartis contributes the largest mean gain to the other datasets, with Tahoe contributing a smaller gain.
Removing VCPI2 or either L1000 source raises OP3 by 0.06--0.09; this target has only 15 VAL compounds.

\paragraph{Structural proximity.}
For each VAL compound, we compare the change from removing Novartis with its maximum Morgan Tanimoto similarity to Novartis TRAIN compounds, first averaging scores over that molecule's evaluation units.
Compounds with a close analog (similarity at least 0.5) lose more on SciPlex ($-0.045$ versus $-0.017$ without), Tahoe ($-0.034$ versus $-0.007$) and L1000 P1 ($-0.011$ versus $+0.001$).
However, compounds without a close analog also lose on OP3 ($-0.064$) and SciPlex, and the within-dataset rank associations are weak (Spearman $-0.15$ to $0.00$).
These molecule-level summaries complement the group--compound averages in the source-removal matrix: structural proximity explains some target differences, but does not account for all gains.

\clearpage
\subsection{Direct Response Copying from Structural Neighbors}
\label{app:neighbor_copying}

A larger training pool provides more close chemical analogs, but their measured responses may depend on dataset and cellular context.
We test direct response copying by comparing a pool of the target dataset's TRAIN compounds with pooled TRAIN compounds from all eight datasets.
For each first-split VAL compound, Morgan Tanimoto similarity ranks the candidates, with ties favoring the target dataset.
Within the target dataset, signatures match cellular context, time and nearest dose, with a molecule-average fallback.
Cross-dataset signatures use the same cell line when available and otherwise the source molecule's mean signature.
The signatures are centered TRAIN responses, aligned by gene identity and scaled by the target-to-source ratio of TRAIN signature RMS.
For each target gene, we average the five nearest candidates that measure it among the retained forty neighbors.

Table~\ref{tab:nearest_neighbor_copying} shows that expanding the pool lowers mean drug contrast from 0.046 to 0.040, improving SciPlex and VCPI1 but reducing the other six datasets.
Thus, broader structural coverage alone does not consistently improve direct response copying under this rule.
Together with the gains from joint molecular training, this supports learning how chemical structure maps to responses across experimental contexts.

\begin{table}[H]
\centering
\caption{\textbf{Direct copying from structural neighbors.} Drug-contrast Pearson on the first split\textquotesingle s VAL compounds, using five TRAIN neighbors per gene. Full/HVG routes are averaged within each dataset; both L1000 routes use the landmark panel.}
\label{tab:nearest_neighbor_copying}
\small
\setlength{\tabcolsep}{11pt}
\begin{tabular}{lrrr}
\toprule
Dataset & Own-dataset pool & Eight-dataset pool & Change \\
\midrule
OP3 & 0.121 & 0.079 & $-0.042$ \\
SciPlex & 0.031 & 0.047 & $+0.016$ \\
VCPI1 & 0.018 & 0.023 & $+0.005$ \\
VCPI2 & 0.021 & 0.015 & $-0.005$ \\
Novartis & 0.049 & 0.041 & $-0.007$ \\
Tahoe & 0.047 & 0.036 & $-0.011$ \\
L1000 P1 & 0.046 & 0.044 & $-0.002$ \\
L1000 P2 & 0.037 & 0.034 & $-0.003$ \\
\midrule
Equal-dataset mean & 0.046 & 0.040 & $-0.006$ \\
\bottomrule
\end{tabular}
\end{table}

\subsection{Complete Hallmark Pathway Results}
\label{app:pathway_results}

\begin{figure}[!p]
\centering
\includegraphics[width=\textwidth]{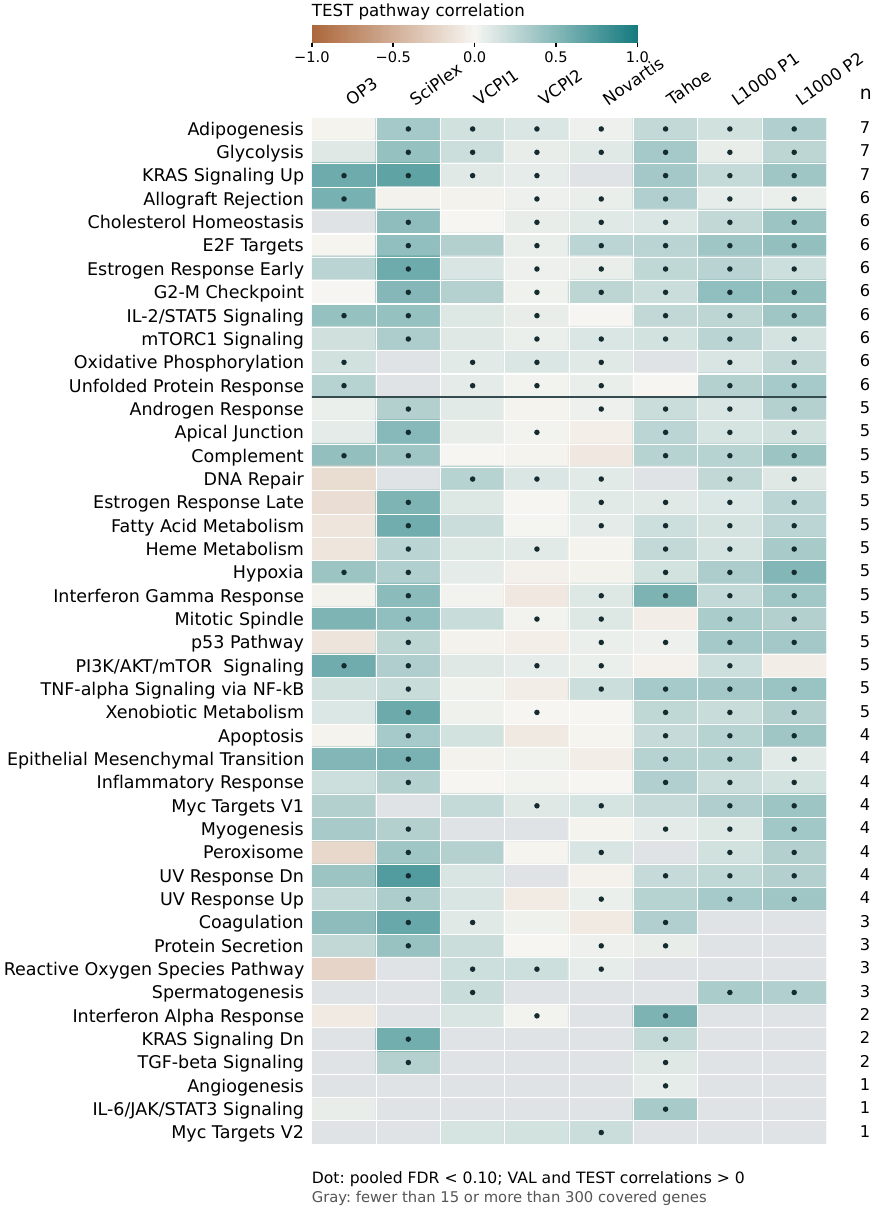}
\caption{\textbf{Complete Hallmark pathway map.} All 44 pathways with eligible gene coverage in at least one dataset. Colors show TEST molecule-level correlation of truth-SD-scaled pathway effects; dots mark pooled VAL/TEST FDR $<0.10$ and positive correlations in both splits. Gray denotes ineligible gene coverage. The right column counts marked datasets, and the separator identifies the twelve pathways displayed in Figure~\ref{fig:interpretability_combined}c. Rows are ordered by this count and then alphabetically.}
\label{fig:pathways_complete}
\end{figure}

We analyze the Enrichr MSigDB Hallmark 2020 gene sets, retaining pathways with 15--300 genes in each dataset's HVG or landmark panel.
Figure~\ref{fig:pathways_complete} includes all 44 pathways eligible in at least one dataset, including cells that do not pass the display criterion.
For each pathway, we scale genes by their true response standard deviation within the analyzed split, average their drug-contrast effects, and then average evaluation units per molecule.
Pearson correlation measures agreement between predicted and observed pathway effects across molecules.
These standardized pathway scores are distinct from the native-gene benchmark correlations.

A one-sided test with 1,000 molecule permutations evaluates the pooled VAL and TEST correlation, followed by Benjamini--Hochberg correction within each dataset's Hallmark library.
A dataset--pathway pair is marked when pooled FDR is below 0.10 and correlations are positive in VAL and TEST separately.
This descriptive criterion identifies 198 pairs: OP3 8, SciPlex 33, VCPI1 9, VCPI2 21, Novartis 25, Tahoe 33, L1000 P1 35 and L1000 P2 34.
Forty-one pathways meet it in at least two datasets and thirty-four in at least four.
The twelve meeting it in at least six datasets appear in the main figure and above the horizontal separator in the complete map.

\end{document}